%% file: arxiv.tex
\documentclass{bmvc2k}

\title{Text-Based Person Search with Limited Data}

\addauthor{Xiao Han}{xiao.han@surrey.ac.uk}{1,2}
\addauthor{Sen He}{sen.he@surrey.ac.uk}{1,2}
\addauthor{Li Zhang}{lizhangfd@fudan.edu.cn}{3}
\addauthor{Tao Xiang}{t.xiang@surrey.ac.uk}{1,2}

\addinstitution{
 CVSSP\\
 University of Surrey\\
}
\addinstitution{
 iFlyTek-Surrey Joint Research Centre\\
on Artificial Intelligence
}
\addinstitution{
 School of Data Science\\
 Fudan University\\
}

\runninghead{Han et al.}{TBPS with Limited Data}

\def\ie{\emph{i.e}\bmvaOneDot}
\def\eg{\emph{e.g}\bmvaOneDot}

\usepackage{listings}
\usepackage{amssymb}
\usepackage{multirow}
\usepackage{multicol}
\usepackage{graphicx}
\usepackage{capt-of}
\usepackage{color}
\definecolor{myblue}{RGB}{0,120,191}
\definecolor{myred}{RGB}{204,64,84}
\definecolor{mygreen}{RGB}{19,138,7}
\newcommand{\first}[1]{\textcolor{myred}{#1}}
\newcommand{\second}[1]{\textcolor{myblue}{#1}}
\newcommand{\revision}[1]{#1}
\newcommand{\statement}[1]{\noindent\textbf{#1}}

\begin{document}

\maketitle

\input{file/0-abstract}
\input{file/1-intro}
\input{file/2-relatedwork}
\input{file/3-method}
\input{file/4-experiments}
\input{file/5-conclusions}
\newpage
\input{file/supp_body}
\newpage
\bibliography{egbib}

\end{document}

%% file: file/0-abstract.tex
\begin{abstract}
Text-based person search (TBPS) aims at retrieving a target person from an image gallery with a descriptive text query.
Solving such a fine-grained cross-modal retrieval task is challenging, which is further hampered by the lack of large-scale datasets.
In this paper, we present a framework with two novel components to handle the problems brought by limited data. 
Firstly, to fully utilize the existing small-scale benchmarking datasets for more discriminative feature learning, we introduce a cross-modal momentum contrastive learning framework to enrich the training data for a given mini-batch. Secondly, we propose to transfer knowledge learned from existing coarse-grained large-scale datasets containing image-text pairs from drastically different problem domains to compensate for the lack of TBPS training data. A transfer learning method is designed so that useful information can be transferred despite the large domain gap.  Armed with these components, our method achieves new state of the art on the CUHK-PEDES dataset with significant improvements over the prior art in terms of Rank-1 and mAP.
Our code is available at \url{https://github.com/BrandonHanx/TextReID}.
\end{abstract}

%% file: file/1-intro.tex
\vspace{2mm}
\section{Introduction}
Text-based person search (TBPS) \cite{li2017person} is the problem of  retrieving a target person from an image gallery with a descriptive text query. 
It is more flexible  compared to image-based person search when the query image is difficult to obtain. 
It has thus gained increasing attention in the research community \cite{ye2021survey,wang2020survey2,lin2021survey3}. 
TBPS has various potential applications such as video surveillance and personal photo album search.

Despite the existing efforts, TBPS is still far from being solved.
One of the reasons is that it is intrinsically challenging as a fine-grained cross-modal retrieval task, where all images belong to the same category, \ie, pedestrian. 
This contrasts with the more widely studied generic image-text retrieval task \cite{chen2020uniter, tan2019lxmert, sun2021lightningdot, li2020oscar}.
The fine-grained nature dictates that more discriminative features must be learned to distinguish visual cues and textual attributes.
This is thus the focus of existing TBPS methods. 
Specifically, prior works \cite{aggarwal2020cmaam,chen2021cmka,gao2021contextual,jing2020pose,sarafianos2019adversarial} typically use a two-stream architecture for fast inference, where both streams are initialized from backbones pre-trained on large-scale unimodal data, \eg, ResNet \cite{he2016resnet} and BERT \cite{devlin2018bert}. 
For the purpose of learning more discriminative features, many methods \cite{aggarwal2020cmaam,gao2021contextual,jing2020pose,wang2020vitaa,zheng2020hierarchical} also take advantage of multi-scale learning, where feature maps with different receptive fields and word/phrase/sentence embeddings are used for the visual and textual stream, respectively.

TBPS is also faced with a second challenge which has been largely ignored, that is, the lack of training data.  
Collecting a large-scale TBPS dataset and annotating it with highly fine-grained text descriptions is tedious and expensive. 
As a result, most existing TBPS dataset is orders of magnitude smaller than those coarse-grained generic image-text pair datasets \cite{chen2020uniter,li2020oscar}.
Having only limited data has clear negative effects on a TBPS model's ability to learn discriminative cross-modal features for fine-grained retrieval. 

Existing methods \cite{aggarwal2020cmaam,chen2021cmka,gao2021contextual,jing2020pose,sarafianos2019adversarial} are ill-equipped to address this limited data problem. 
More specifically, most of their learning objectives require the training data to be organized into positive and negative pairs.
However, previous works construct negative pairs merely from a certain mini-batch, which does not make full use of the available TBPS data. 
Pre-training on larger image-text pair datasets is also an obvious option to compensate for the lack of training data.  
Nevertheless, the visual and textual streams in previous works are initialized from models that are separately pre-trained on unimodal data rather than image-text pairs. 
Information useful for cross-domain matching is thus not exploited. 
Cross-modal pre-training may have been attempted. 
However, as shown in this work, without a careful design, a naive pre-training then fine-tuning strategy would lead to negative transfer.   

To overcome the learning with limited data problem, in this work, we propose a framework with two novel components for TBPS. 
Firstly, to fully utilize the existing small-scale benchmarking datasets for more discriminative feature learning, we introduce a cross-modal momentum contrastive learning (or CM-MoCo) framework to enrich the training data for a given mini-batch.
CM-MoCo decouples the number of negative pairs with the mini-batch size to obtain more negative cross-modal counterparts for each image or description. 
To implement such a framework, in addition to the two gradient-updated encoders (query encoders), we introduce another two momentum-updated encoders (key encoders) for two modalities and maintain three different queues to store visual features, textual features, and identities from previous batches. 
Further, a contrastive loss is formulated in a cross-modal manner, which treats the features from query encoders, key encoders and queues as anchors, positive samples and negative samples, respectively. 
Secondly, a cross-modal transfer learning method is proposed to benefit from large-scale coarse-grained image-text pair datasets.
Instead of the commonly used pre-training + fine-tuning strategy, we propose to freeze the text encoder of the pre-trained model to embed each word and then adopt one bidirectional GRU layer (Bi-GRU) \cite{cho2014gru} to contextualize words.
Empirically, this transfer learning strategy can effectively prevent the negative transfer suffered by the naive full model transfer strategy (See Table~\ref{ablation} for experimental results).   

The main contributions of this paper are: 
(1) A novel cross-modal momentum contrastive learning framework is proposed to better utilize the existing small-scale TBPS datasets. (2) To effectively transfer the knowledge learned from large-scale generic image-text pairs, we propose to perform cross-modal pre-training, but for the text modality, only word embedding is transferred. 
(3) Extensive experiments are conducted to show that our proposed framework outperforms existing methods on CUHK-PEDES \cite{li2017person} by large margins.

%% file: file/2-relatedwork.tex
\section{Related work}
\subsection{Text-based person search}
\citet{li2017person} first propose TBPS with a challenging dataset CUHK-PEDES and a baseline built upon a recurrent neural network with gated neural attention.
Following this, many single-scale methods are proposed for better investigating the intra- and inter-modal fine-grained differences with the help of instance loss \cite{zheng2020dual}, cross-modal projection loss \cite{zhang2018cmpc}, adversarial loss \cite{sarafianos2019adversarial} and cross-modal knowledge adaption \cite{chen2021cmka}.

Besides, some multi-scale methods are proposed to learn the semantic relevance between specific image regions and phrases/words in descriptions. 
Many works implement such architecture by making local image features attend to corresponding noun phrases and words through a variety of attention mechanisms \cite{niu2020mia,zheng2020hierarchical,gao2021contextual,farooq2021axm,wang2021mgel}.
Additionally, some works adopt side information to help align two modalities, \eg, pose information \cite{jing2020pose}, semantic segmentation maps \cite{wang2020vitaa} and attribute labels \cite{aggarwal2020cmaam}. 
Most of these multi-scale architectures merely use global features during inference, because calculating the similarity between local features increases both the inference time and the offline features storage space.

For fast inference speed and less memory consumption, neither multi-scale architecture nor side information is involved in our method.
Instead of manually designing more complicated network architectures or collecting more side information, from a more general perspective, we focus on a much more practical and under-studied problem in TBPS, \ie, the scarcity of data. \revision{Being orthogonal to all previous methods, our method can be easily extended or integrated. Many solid experiments show our method is comparable and even better with other more complex methods.}

\vspace*{-3mm}
\subsection{Contrastive learning}
The basic idea of contrastive learning is to map the original data into a latent feature space where the similarities between positive/negative pairs are maximized/minimized \cite{hadsell2006contrastive}. 
The instance discrimination is the most prevalent pretext task, whose positive pairs consist of two augmented views of the same instance, and the other pairs are defined to be negative. 
MoCo \cite{he2020moco} and SimCLR \cite{chen2020simclr} suggest that large quantities of data pairs are crucial to the performance of contrastive learning. 
Most recently, BYOL \cite{grill2020byol} and SimSiam \cite{chen2020simsiam} prove that negative pairs are unnecessary and the invariant observation of the same concept matters.

In this work, to fully exploit the available annotated dataset, we apply momentum contrastive learning to TBPS. 
There are two differences between our work and the classic instance discrimination contrastive learning framework: 
(1) Our task is identity-level rather than instance-level, because each identity has more than one image and description in the dataset. 
(2) In our task, the similarity is measured in a cross-modal manner.

\vspace*{-3mm}
\subsection{Vision-language pre-training}
With the advent of Transformer \cite{vaswani2017transformer} and BERT \cite{devlin2018bert}, there has been a surging interest in applying self-supervised learning to multimodal tasks. This is usually done by pre-training on large-scale generic image-text pairs and then fine-tuning on downstream tasks.
ViLBERT \cite{lu2019vilbert} and LXMERT \cite{tan2019lxmert} introduce the two-stream architecture, where two Transformers are applied to images and text independently followed by another Transformer for cross-modal fusion. 
In addition to that, many works \cite{su2019vlbert, chen2020uniter,li2020oscar,kim2021vilt} adopt the single-stream architecture and achieve much better performance. 
In such an architecture, a single Transformer is applied to both images and text.

Although single stream models have achieved great success, its crucial component, cross-modal attention between two modalities, triggers the inevitable latency and significant computation during training and inference. 
To tackle this problem, CLIP \cite{radford2021clip} and BriVL \cite{huo2021wenlan} utilise larger datasets, larger batch size, and contrastive learning on the basis of two-stream architecture. 
LightningDOT \cite{sun2021lightningdot} adopts a faster two-stream model as the main inference model and another stronger single-stream model as the re-ranker, achieving a satisfactory balance between accuracy and efficiency.

In this work, we incorporate pre-trained two-stream models into TBPS. 
We study how to effectively transfer the knowledge pre-trained on large-scale coarse-grained image-text pairs for fine-grained TBPS in spite of the big domain gap between them.

%% file: file/3-method.tex
\vspace*{-3mm}
\section{Methodology}

\input{figure/architecture}
Given a text query $t$, the goal of TBPS is to retrieve an image $v$ that best matches the content in $t$ from a gallery. 
The retrieval is successful if $t$ and $v$ share the same identity.

Our proposed framework is illustrated in Figure~\ref{overall_arch}. 
It consists of two query encoders $f_{q}^{V}$ and $f_{q}^{T}$ (Visual and Textual Q-Encoder) along with two key encoders $f_{k}^{V}$, $f_{k}^{T}$ (Visual and Textual K-Encoder), parameterized by $\theta_{q}^{V}$, $\theta_{q}^{T}$, $\theta_{k}^{V}$, $\theta_{k}^{T}$, respectively. 
During training, both the query and key encoders are used to process the input from its own modality. 
The outputs of the key encoders are pushed into queues which are used to construct negative pairs for contrastive learning. 
During inference, only two query encoders are used for feature extraction. 
The retrieval is done by first computing the cosine similarity between the query feature and the offline extracted features of all candidates in the gallery, and then selecting the candidate that has the highest similarity score.

In the following sections, we will first introduce our proposed cross-modal momentum contrastive learning pipeline (CM-MoCo) in Section \ref{vta}, and then explain how we effectively transfer the knowledge learned from large-scale generic image-text pairs in Section \ref{transfer}.

\vspace*{-3mm}
\subsection{Learning from limited TBPS data}
\label{vta}
\statement{Cross-modal momentum contrastive learning.}
One of the limitations to learn more discriminative features in previous work is caused by the limited negative pairs during the training stage.
Note that MoCo \cite{he2020moco} provides a mechanism of building dynamic queues decoupled with batch size, which makes it possible to learn from more negative samples beyond a certain batch. 
Inspired by this, we propose cross-modal momentum contrastive learning to make the best use of current TBPS data. 

Concretely, given a batch of person images $V=\{v_1,\cdots,v_B\}$, a batch of descriptions $T=\{t_1,\cdots,t_B\}$ and their identities $ID=\{id_1,\cdots,id_B\}$, we feed $V$ and $T$ into their corresponding query encoder and key encoder to obtain their normalized features:
\begin{equation}
    \mathbf{V}^{q} = f_{q}^{V}(V),\ \mathbf{V}^{k} = f_{k}^{V}(V),\
    \mathbf{T}^{q} = f_{q}^{T}(T),\
    \mathbf{T}^{k} = f_{k}^{T}(T),
\end{equation}
where $\mathbf{V}^q,\mathbf{V}^k,\mathbf{T}^q,\text{and} \ \mathbf{T}^k \in \mathbb{R}^{B \times D}$. $\mathbf{V}^k$, $\mathbf{T}^k$ and $ID$ will be pushed into three queues, \ie, visual queue ($Q^V$), textual queue ($Q^T$) and identity queue ($Q^{ID}$) for negative pair construction. 

To learn discriminative cross-modal features, for each image query $v_i$, we define the cross-modal contrastive loss among its query feature $\mathbf{V}_i^q$ (as anchor), its corresponding textual key feature $\mathbf{T}_i^k$ (as positive key) and the keys stored in textual queue $\widetilde{Q}^T$ (as negative keys), where $\widetilde{Q}^T=\{\mathbf{n}|\mathbf{n}\in Q^T \wedge Q^{ID}(\mathbf{n})\notin ID\}$ indicating its identities are not in the current batch. 
In the meanwhile, we also apply the cross-modal contrastive learning in a symmetrical way when regarding each description $t_i$ as a query. 
The overall cross-modal contrastive loss $\mathcal{L}_{cmc}$ is computed as Equation \ref{cmc}, where $\tau_c$ denotes the tuneable temperature.
\begin{equation}
\mathcal{L}_{cmc}=-\sum_{i=1}^{B} \log\left[\frac{e^{\left(\mathbf{V}_i^q \mathbf{T}_i^k / \tau_c\right)}}{e^{ \left(\mathbf{V}_i^q \mathbf{T}_i^k / \tau_c\right)}+\displaystyle\sum\limits_{\mathbf{n}\in \widetilde{Q}^T} e^{\left(\mathbf{V}_i^q \mathbf{n} / \tau_c\right)}}\right] -\sum_{i=1}^{B} \log\left[\frac{e^{\left(\mathbf{T}_i^q\mathbf{V}_i^k / \tau_c\right)}}{e^{ \left(\mathbf{T}_i^q  \mathbf{V}_i^k / \tau_c\right)}+\displaystyle\sum\limits_{\mathbf{n}\in \widetilde{Q}^V} e^{\left(\mathbf{T}_i^q \mathbf{n} / \tau_c\right)}}\right].
\label{cmc}
\end{equation}

After calculating the cross-modal contrastive loss, two query encoders are updated by the back propagation gradients. 
Following MoCo \cite{he2020moco}, the parameters of two key encoders, $\theta_{k}^{V}$ and $\theta_{k}^{T}$, are updated by the rule given in Equation \ref{moco_update}, where $m$ is a momentum parameter.
\begin{equation}
\theta_{k}^{V}=m \cdot \theta_{k}^{V}+(1-m) \cdot \theta^{V}_{q} ,\ \theta_{k}^{T}=m \cdot \theta_{k}^{T}+(1-m) \cdot \theta^{T}_{q}.
\label{moco_update}
\end{equation}

With our proposed $\mathcal{L}_{cmc}$, each query sample is compared with a large number of negative key samples. 
It thus allows the model to learn more discriminative features for TBPS. 
To better understand CM-MoCo, please refer to the pseudocode in the supplementary material.

\statement{Joint training.} Following previous works, we also incorporate widely used alignment loss $\mathcal{L}_{align}$ \cite{wang2020vitaa} and identity loss $\mathcal{L}_{id}$ \cite{luo2019strong,zheng2020dual} into our end-to-end training pipeline (See details in the supplementary material). 
The overall loss $\mathcal{L}$ is the summation of the three losses:
\begin{equation}
\label{all_loss}
\mathcal{L}=\mathcal{L}_{cmc}+\mathcal{L}_{align}+\mathcal{L}_{id}.
\end{equation}

\statement{Post-processing.} In the inference stage, we adopt the cross-modal $k$-reciprocal rerank algorithm \cite{gao2021contextual} to further improve the performance. 
The pair-wise rerank similarity is calculated by Jaccard Distance of $k$-nearest unimodal neighbors and $k$-nearest cross-modal neighbors, and then added to the original cosine similarity (See details in the supplementary material).

\vspace*{-3mm}
\subsection{Transferring knowledge from generic image-text pairs}
\label{transfer}

\input{figure/text_encoder}
A conventional way for TBPS is to initialize our visual and textual encoders with backbones separately pre-trained on unimodal data, \eg, ResNet50 \cite{he2016resnet} pre-trained on ImageNet \cite{deng2009imagenet} and BERT \cite{su2019vlbert} pre-trained on large corpora.
However, this initialization brings a significant heterogeneous gap which is difficult to be bridged with the current limited data of TBPS. 
To tackle this issue, a straightforward way is to initialize our encoders pre-trained on large-scale generic image-text pairs, \eg, MSCOCO captions \cite{lin2014coco}, Flickr30k \cite{plummer2015flickr30k} and WIT \cite{radford2021clip}, and then fine-tune the whole model for TBPS. 

Unexpectedly, we empirically find this intuitive transfer strategy yields poor performance. 
This negative transfer is likely to be caused by the domain gap between the TBPS domain and that of the generic datasets. 
\revision{Such domain gap especially exists in the textual side, even under the unimodal pre-training scenario. Due to the highly fine-grained specialty of TBPS, the description sentences in TBPS are much longer than those in the generic data, and every word matters. However, text backbones pre-trained on generic data are likely to get more coarse-grained information and neglect some detailed words, thus negative transfer happens (See more discussions in the supplementary material). Nonetheless, we believe pre-training on generic data can still offer us more meaningful embeddings for each word because of largr-scale contrastive learning.}

To address this problem, we propose a transfer learning strategy with three alterations for the text stream while leaving the visual stream unchanged. 
Concretely, as illustrated in Figure \ref{wt_embedding}, we take CLIP Text Encoder (CLIP-TE) \cite{radford2021clip} pre-trained on WIT as an example to demonstrate our alterations. 
Firstly, one of our alterations is feeding the whole sentence into CLIP-TE in a word-by-word manner to obtain word-type embeddings. 
For each word, its word-type embedding is represented by the \texttt{[EOS]} token from the last layer of CLIP-TE. 
Secondly, CLIP-TE is frozen in the whole training stage. 
In the actual implementation, an offline dictionary storing all word embeddings is computed in advance, and the frozen CLIP-TE is removed from the training process. 
Thirdly, to compensate the lack of sentence-level semantic information, we append a Bi-GRU \cite{cho2014gru} followed by max pooling to contextualize all word-type embeddings in a sentence. 
Empirically, as shown in Table~\ref{ablation}, the model with our proposed transfer learning strategy yields a significant performance boost. 
This strategy allows the textual stream to effectively transfer the knowledge learned from large-scale generic image-text pairs.

%% file: figure/architecture.tex
\begin{figure}[t]
\centering
\includegraphics[width=\linewidth]{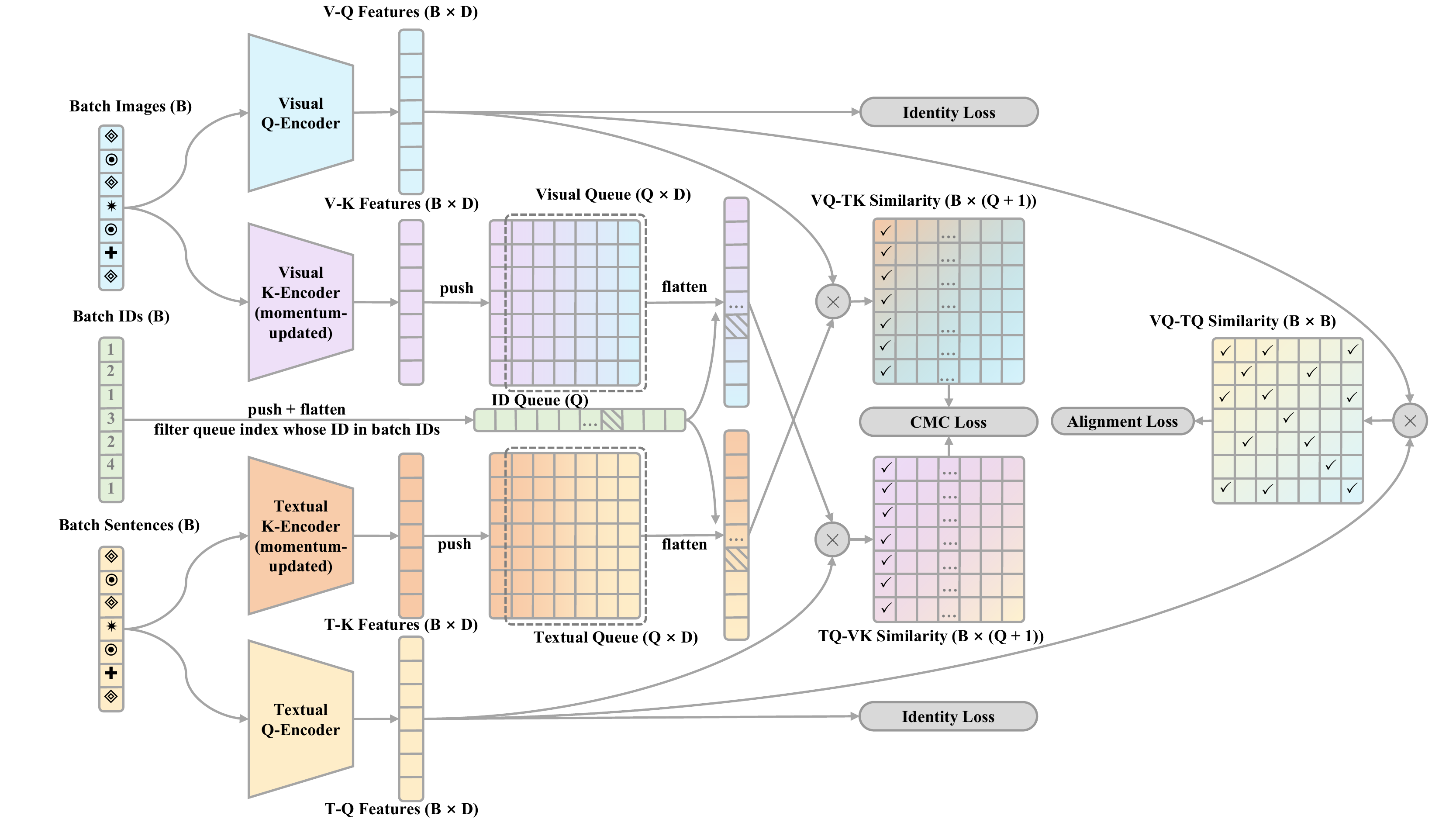}
\vspace*{-6mm}
\caption{Illustrative diagram of our proposed architecture. The two Q-Encoders are gradient-updated while two K-Encoders are initialized from Q-Encoders and momentum-updated. The cross-modal contrastive (CMC) loss, alignment loss and identity loss are employed during training. The whole model is trained in an end-to-end manner and only two Q-Encoders are used for inference.}
\label{overall_arch}
\vspace*{-3mm}
\end{figure}

%% file: figure/text_encoder.tex
\begin{figure}[t]
\centering
\includegraphics[width=0.7\linewidth]{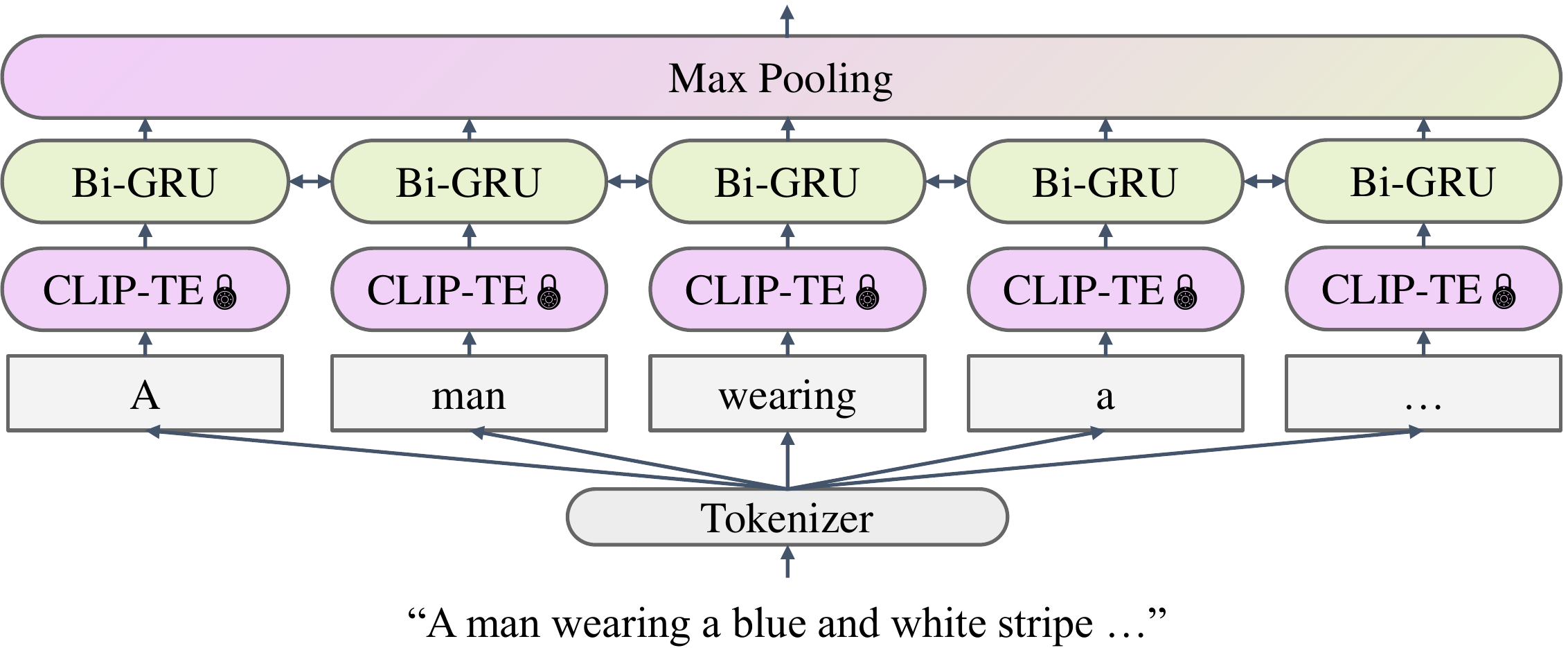}
\caption{\revision{Illustrative diagram of our textual stream. We first tokenize a sentence into words, and then independently feed these words into a pre-trained frozen text backbone (\eg, CLIP text encoder) to get word-type embeddings. On top of that, a Bi-GRU with a max pooling layer is used to contextualize all word embeddings.}}
\label{wt_embedding}
\vspace*{-3mm}
\end{figure}

%% file: file/4-experiments.tex
\vspace*{-3mm}
\section{Experiments}
\subsection{Experimental setup}
\statement{Dataset.} We conduct experiments on the CUHK-PEDES dataset \cite{li2017person}, which is currently the only benchmark for TBPS. 
It contains 40,206 images of 13,003 different people, where each image has two descriptive sentences annotated by different people. 
As per standard, the dataset is split into 11,003 identities with 34,054 images in the training set, 1,000 identities with 3,074 images in the test set, and the remaining for the validation set. 
The average length of all sentences is 23 and the vocabulary size is 9408.

\statement{Implementation details.} During training, we use random horizontally flipping, random crop with padding, and random erasing \cite{luo2019strong} as image data augmentation methods. 
Following previous works \cite{gao2021contextual,zheng2020hierarchical,wang2020vitaa}, the stride of the last block in the ResNet are set to 1 to increase the resolution of the final feature map. 
The feature dimension $D$ for both modalities is set to 256.  
All images are resized to $384 \times 128$. 
Our model is trained with Adam optimizer \cite{kingma2014adam} for 80 epochs with an initial learning rate $1 \times 10^{-4}$, which is decayed by a factor 0.1 at the $40^{th}$ epoch and $70^{th}$ epoch, respectively. 
At the beginning, we spend 5 warm-up epochs linearly increasing the learning rate from $1 \times 10^{-5}$ to $1 \times 10^{-4}$. 
Each mini-batch has 128 image-text pairs with 4 images/sentences for each identity. 
$\tau_c$ in $\mathcal{L}_{cmc}$ is set to $0.07$ and the momentum $m$ is set to $0.999$. 
Following NAFS \cite{gao2021contextual}, the number of nearest neighbors $k$ used in rerank is set to 5.
All experiments are conducted on one V100 GPU with Pytorch \cite{paszke2019pytorch}. 

\statement{Evaluation protocol.} As per standard, we evaluate our model in a bi-directional manner, where the performance is measured by Rank-K (K=1, 5, 10). \revision{Specifically, given a text/image query, Rank-K reports the percentage of successful searches among all searches, where each successful search retrieves at least one corresponding person correctly among the top K results.}
In addition, for a comprehensive evaluation, we also adopt the mean Average Precision (mAP) of all queries \cite{zheng2020hierarchical,farooq2021axm,wang2021mgel} as another retrieval criterion. \revision{Empirically, Rank-K can reflect models' accuracy on the first few retrieval results while mAP puts more emphasis on the order of the entire retrieval sequence predicted by the models.}

\vspace*{-3mm}
\subsection{Main results}
\input{table/sota}
We compare our method with most published works on TBPS. 
We integrate the proposed CM-MoCo and transfer the knowledge from generic data to finalize our model. 
For a comprehensive comparison, we instantiate our method with both CLIP ResNet50 and CLIP ResNet101.
\revision{Please notice that we cannot reach 100\% fair comparisons with all other methods, because the details of the implementations vary. For example, in the pursuit of a more general and efficient structure, being designed in a single-scale architecture with only 256 feature dimensions leads our method into a natural inferior position.}

As shown in Table~\ref{sota}, our models' performances are comparable and even better with other complicated methods in terms of all metrics no matter whether rerank post-processing is used.
Specifically, for the most important metric Rank-1, our model (ResNet101) gains approximately 9\%/1\% absolute improvement over the previous single-/multi-scale state-of-the-art method. 
\revision{For the image to text retrieval, our method outperforms all others by a large margin, indicating our method's superiority in aligning two modalities. The highest mAP also demonstrates the entire retrieval sequence predicted by our models has a top-quality order. In addition to the satisfactory performance, our method also has the merits over the training efficiency, retrieval speed and offline features storage against previous state-of-the-art methods (See detailed comparisons in the supplementary material).}

\subsection{Ablation studies}
\input{table/component_ablation}
\input{table/other_ablation}
We evaluate the effectiveness of our framework by introducing our proposed CM-MoCo in Section \ref{vta} and transfer learning from cross-modal pre-training in Section \ref{transfer}.
We adopt a baseline model trained without our proposed $\mathcal{L}_{cmc}$ and transfer learning, where both visual and textual backbones are initialized with the models separately pre-trained on unimodal data, \ie, ResNet101 and BERT.  
To illustrate the power of cross-modal pre-training with generic image-text pairs in TBPS, in addition to adopting CLIP pre-trained on the unreleased huge-scale WIT \cite{radford2021clip} (about 400 million pairs), we pre-train another model on a relative smaller dataset, MSCOCO captions \cite{lin2014coco} (about 0.57 million pairs), to reach a comprehensive comparison (See more details for pre-training settings in the supplementary material). 

We can draw the following conclusions from the Table \ref{ablation}: 
(1) Without our transfer strategy, no matter whether and how many generic image-text pairs are used for pre-training, directly fine-tuning the pre-trained model results in a negative transfer. \revision{This transfer strategy can effectively alleviate the domain gap coming from the textual side}. More specifically, it leads to at least $6.66\%$ performance improvement.
(2) \revision{When the domain gap is well resolved, cross-modal pre-training yields better performance than unimodal pre-training.} This is even more significant when the image-text pairs dataset for pre-training is scaled up, \ie, from MSCOCO to WIT. 
(3) The proposed CM-MoCo yields consistent improvement for all models. It further boosts the performance with $1.5\%$ improvement in average. 

\newpage
\statement{Transfer learning strategy.} In this part, we validate the design of our transfer learning strategy proposed in Section \ref{transfer}. 
There are three proposed alterations, \ie, GRU, fixed textual encoder and non-contextualized word embedding. 
Table \ref{ablation_text} demonstrates the effectiveness of each alteration. 
It is clear that appending a GRU to the pre-trained textual encoder directly allows the textual stream to learn fine-grained textual information on the basis of the previously learned coarse-grained one. 
Further, fixed textual encoder and non-contextualized word embedding boost the performance with about 4\% and 1\%, respectively.
\revision{This confirms our conjecture that large-scale cross-modal pre-training can provide more meaningful word embeddings and image features, while the domain gap problem is needed to be addressed carefully.}
In summary, this strategy allows us to effectively transfer the knowledge from cross-modal pre-training. 

\statement{Queue sizes.} We also evaluate the effect of the queue sizes that is used to store keys and construct negative pairs for CM-MoCo. 
From the results reported in Table \ref{ablation_moco}, it is obvious that a large queue size (1024 and 2048) in CM-MoCo improves the performance. 
However, further increasing this queue size yields worse performance. 
This phenomenon is probably due to the limited data in the TBPS dataset.
A queue size which is too large for the entire dataset may store too many obsolete keys, which will mislead the learning direction. 
It is therefore detrimental for further cross-modal contrastive learning.

\vspace*{-3mm}
\subsection{Qualitative results} Figure \ref{vis_result} visualizes some typical retrieval results (See more successful results and failure cases in the supplementary material). 
Concretely, Figure \ref{vis_1} is a retrieval result with a detailed text query, our method successfully retrieves the target image with a high similarity score.
Figure \ref{vis_2} and \ref{vis_3} show the results of two different text queries for the same target image.
The second query describes the lady's coat and shoes as "black clothes" and "black shoes", while the third query is more fine-grained with "thigh-level black coat" and "tall black high heels" and thus gets a much better result. We can draw two conclusions from this figure:
(1) Our method appears to be very effective in distinguishing fine-grained details in a given text query.
(2) When the text query has ambiguity, our method can still give reasonable results.
\revision{(3) The result sequence retrieved by our method conforms to human intuition.}
\input{figure/retrieval_vis}

%% file: table/sota.tex
\begin{table}[t]
\begin{center}
\resizebox{\textwidth}{!}{
\begin{tabular}{|c|c|c|c|c|c|c|c|c|c|c|}
\hline
\multirow{2}{*}{\textbf{Method}} & \multirow{2}{*}{\textbf{Arch.}} & \multirow{2}{*}{\textbf{Dim.}} & \multicolumn{4}{c|}{\textbf{Text to Image w/o Rerank}}             & \multicolumn{4}{c|}{\textbf{Image to Text w/o Rerank}}             \\ \cline{4-11} 
                                 &                      &                 & \textbf{Rank-1} & \textbf{Rank-5} & \textbf{Rank-10} & \textbf{mAP} & \textbf{Rank-1} & \textbf{Rank-5} & \textbf{Rank-10} & \textbf{mAP} \\ \hline \hline
\rowcolor[RGB]{230,230,230}
GNA-RNN \cite{li2017person}                          & S   & 512                        & 19.05           & -               & 53.64            & -            & -               & -               & -                & -            \\ \hline
\rowcolor[RGB]{230,230,230}
Dual Path \cite{zheng2020dual}                        & S & 2048                          & 44.40           & 66.26           & 75.07            & -            & -               & -               & -                & -            \\ \hline
\rowcolor[RGB]{230,230,230}
CMPM/C$\dagger$ \cite{zhang2018cmpc}             & S  & 512                            & 49.37           & 71.69           & 79.27            & -        & 60.96           & 84.42           & 90.83            & -        \\ \hline
MIA \cite{niu2020mia}                             & M  & 1024                              & 53.10           & 75.00           & 82.90            & -            & -               & -               & -                & -            \\ \hline
PMA \cite{jing2020pose}                              & M & 1024                               & 54.12           & 75.45           & 82.97            & -            & -               & -               & -                & -            \\ \hline
\rowcolor[RGB]{230,230,230}
TIMAM$\dagger$ \cite{sarafianos2019adversarial}                          & S  & 512                             & 54.51           & 77.56           & 84.78            & -        & 67.40           & 88.65           & 93.91            & -        \\ \hline
\rowcolor[RGB]{230,230,230}
CKMA \cite{chen2021cmka}                             & S   & 512                             & 54.69           & 73.65           & 81.86            & -            & -               & -               & -                & -            \\ \hline
ViTAA$\ddagger$ \cite{wang2020vitaa}                           & M & 256                              & 54.92           & 75.18           & 82.90            & 51.60        & 65.71           & 88.68           & 93.75            & 45.75        \\ \hline
CMAAM \cite{aggarwal2020cmaam}                            & M  & 512                             & 56.68           & 77.18           & 84.86            & -            & -               & -               & -                & -            \\ \hline
HGAN$\dagger$ \cite{zheng2020hierarchical}                         & M   & 512                              & 59.00           & 79.49           & 86.62            & -        & 71.16      & 90.05           & 95.06            & -        \\ \hline
NAFS (G)$\ddagger$ \cite{gao2021contextual}                            & M   & 768                           & 59.36           & 79.13          & 86.00            & 54.07            & 71.89               & 90.99               & 95.28                & 50.16            \\ \hline
MGEL \cite{wang2021mgel}                            & M & 512                             & 60.27           & 80.01          & 86.74            & -            & 71.87               & 91.38               & 95.42                & -            \\ \hline
AXM-Net \cite{farooq2021axm}                          & M & 512                             & 61.90           & 79.41           & 85.75            & 57.38       & -               & -               & -                & -            \\ \hline
TIPCB$\ddagger$ \cite{chen2021tipcb}                          & M  & 2048                            & \second{63.63}           & \first{82.82}           & \first{89.01}            & 56.78       & 73.55               & 92.26               & 96.03                & 51.78            \\ \hline
\rowcolor[RGB]{230,230,230}
Ours (ResNet50)                             & S  & 256                                   & 61.65           & 80.98           & 86.78            & \second{58.29}        & \second{75.96}           & \second{93.40}           & \second{96.55}            & \second{55.05}        \\ \hline
\rowcolor[RGB]{230,230,230}
Ours (ResNet101)                            & S  & 256                                   & \first{64.08}           & \second{81.73}           & \second{88.19}            & \first{60.08}        & \first{78.99}           & \first{95.02}           & \first{97.17}            & \first{56.78}        \\ \hline \hline
\multirow{2}{*}{\textbf{Method}} & \multirow{2}{*}{\textbf{Arch.}} & \multirow{2}{*}{\textbf{Dim.}} & \multicolumn{4}{c|}{\textbf{Text to Image w/ Rerank}}              & \multicolumn{4}{c|}{\textbf{Image to Text w/ Rerank}}              \\ \cline{4-11} 
            &                     &                                       & \textbf{Rank-1} & \textbf{Rank-5} & \textbf{Rank-10} & \textbf{mAP} & \textbf{Rank-1} & \textbf{Rank-5} & \textbf{Rank-10} & \textbf{mAP} \\ \hline \hline
ViTAA$\ddagger$ \cite{wang2020vitaa}                           & M & 256                              & 54.92           & 74.77           & 82.49            & 52.60        & 66.17           & 88.61           & 93.56            & 46.39        \\ \hline
NAFS (G)$\ddagger$ \cite{gao2021contextual}                           & M  & 768                            & 59.62           & 78.90           & 85.72            & 55.02            & 72.67           & 90.92          & 95.12            & 50.92            \\ \hline
TIPCB$\ddagger$ \cite{chen2021tipcb}                           & M    & 2048                          & \second{63.37}           & \first{81.56}           & \second{87.57}            & \second{60.02}            & 74.04           & 92.06          & 95.61            & 53.78            \\ \hline
\rowcolor[RGB]{230,230,230}
Ours (ResNet50)                             & S     & 256                                & 61.94           & 80.52           & 86.45            & 59.45        & \second{76.26}           & \second{93.46}           & \second{96.58}            & \second{55.67}        \\ \hline
\rowcolor[RGB]{230,230,230}
Ours (ResNet101)                             & S   & 256                                  & \first{64.40}           & \second{81.27}           & \first{87.96}            & \first{61.19}        & \first{78.99}           & \first{95.02}           & \first{97.23}            & \first{57.31}        \\ \hline
\end{tabular}
}
\end{center}
\caption{Comparisons with previous methods on the CUHK-PEDES. \revision{Only global features are used during inference for our reproduced NAFS \cite{gao2021contextual}. "Arch."/"Dim." is the abbreviation for architecture/feature dimension. S/M stands for the methods designed in single-/multi-scale architecture, and all single-scale methods are highlighted with \textcolor{gray}{gray} background.} $\dagger$ stands for the results from HGAN \cite{zheng2020hierarchical}. $\ddagger$ stands for the results reproduced with public codes/checkpoints released by their authors. Overall $1^{st}/2^{nd}$ best in \first{red}/\second{blue}.}
\label{sota}
\end{table}

%% file: table/component_ablation.tex
\begin{table}[t]
\begin{center}
\resizebox{\textwidth}{!}{
\begin{tabular}{|c|c|c|c|c|c|c|c|}
\hline
\textbf{Visual Backbone} & \textbf{Textual Backbone}  & \textbf{Paired Data} & \textbf{Transfer Strategy} & \textbf{CM-MoCo} & \textbf{Rank-1} & \textbf{Rank-5} & \textbf{Rank-10} \\ \hline \hline
ResNet101                & BERT                       & -                              &                      &                  & 51.95           & 71.30           & 79.78            \\ \hline
ResNet101                & BERT                       & -                              &  \checkmark          &                  & 58.24           & 78.64           & 85.62            \\ \hline
ResNet101                & BERT                       & -                              &  \checkmark          &    \checkmark    & 59.62           & 79.13           & 86.08            \\ \hline
ResNet101                & BERT                       & MSCOCO \cite{lin2014coco}      &                      &                  & 53.30           & 72.77           & 80.02            \\ \hline
ResNet101                & BERT                       & MSCOCO \cite{lin2014coco}      &  \checkmark          &                  & 59.96           &   79.74           & 86.83            \\ \hline
ResNet101                & BERT                       & MSCOCO \cite{lin2014coco}      &  \checkmark          &    \checkmark    & 60.79           &   79.58         & \second{87.35}         \\ \hline
CLIP ResNet101           & CLIP-TE                    & WIT \cite{radford2021clip}     &                      &                  & 0.15            & 0.76            & 1.23             \\ \hline
CLIP ResNet101           & CLIP-TE                    & WIT \cite{radford2021clip}     &  \checkmark          &                  & \second{62.52}  & \second{80.57}  & 87.12   \\ \hline 
CLIP ResNet101           & CLIP-TE                    & WIT \cite{radford2021clip}     &  \checkmark          &    \checkmark    & \first{64.08}   & \first{81.73}   & \first{88.19}    \\ \hline
\end{tabular}
}
\end{center}
\caption{Ablation experimental results for proposed components. Only the results without rerank for text-to-image task are reported. Paired data denotes the dataset used for pre-training. Transfer strategy denotes the proposed strategy mentioned in Section \ref{transfer}.}
\label{ablation}
\end{table}

%% file: table/other_ablation.tex
\begin{table}[t]
\begin{minipage}{0.525\textwidth}
\resizebox{\textwidth}{!}{
\begin{tabular}{|c|c|c|l|c|c|c|c|}
\hline
\textbf{Embed. Type}    & \textbf{GRU} & \textbf{Fixed} & \textbf{Rank-1} & \textbf{Rank-5} & \textbf{Rank-10} \\ \hline \hline
Contextualized          &              &                & 0.15            & 0.76            & 1.23             \\ \hline
Contextualized          & \checkmark   &                & 57.23           & 75.39           & 82.39            \\ \hline
Contextualized          & \checkmark   & \checkmark     & \second{61.39}  & \second{79.73}  & \second{86.84}   \\ \hline
Word-type               & \checkmark   & \checkmark     & \first{62.52}   & \first{80.57}   & \first{87.12}    \\ \hline
\end{tabular}
}
\vspace*{-1mm}
\captionof{table}{Results for different text encoders.}
\label{ablation_text}
\end{minipage}
\begin{minipage}{0.475\textwidth}
\resizebox{\textwidth}{!}{
\begin{tabular}{|c|c|c|c|c|c|}
\hline
\textbf{CM-MoCo} & \textbf{Queue Size} & \textbf{Rank-1} & \textbf{Rank-5} & \textbf{Rank-10} \\ \hline \hline
              & 0                   & 62.52           & 80.57           & 87.12            \\ \hline
\checkmark    & 1024                & \second{63.52}  & \first{81.77}   & \first{88.52}    \\ \hline
\checkmark    & 2048                & \first{64.08}   & \second{81.73}  & \second{88.19}   \\ \hline
\checkmark    & 4096                & 63.06           & 81.50           & 87.69            \\ \hline
\end{tabular}
}
\vspace*{-1mm}
\captionof{table}[h]{Results for different queue sizes.}
\label{ablation_moco}
\end{minipage}
\vspace*{-5mm}
\end{table}

%% file: figure/retrieval_vis.tex
\begin{figure}[t]
\centering
\subfigure[\underline{He} is wearing \underline{grey pants}, a \underline{dark long sleeved sweater}, and a \underline{light collared shirt underneath}. He is carrying a \underline{black backpack} on two shoulders.]{
\includegraphics[width=\linewidth]{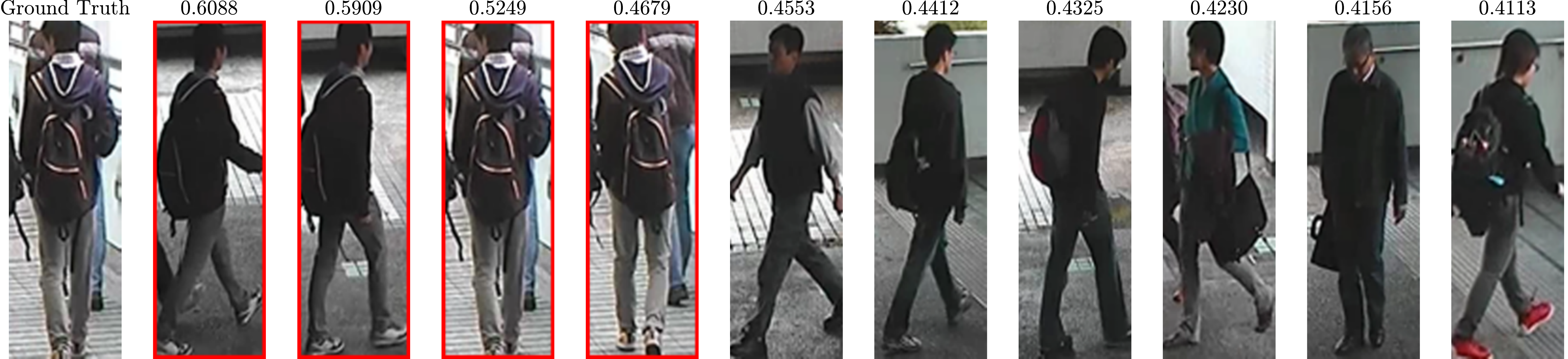}
\label{vis_1}
}
\subfigure[The \underline{woman} with \underline{dark hair} has all \underline{black clothes and shoes} with a \underline{white handbag}.]{
\includegraphics[width=\linewidth]{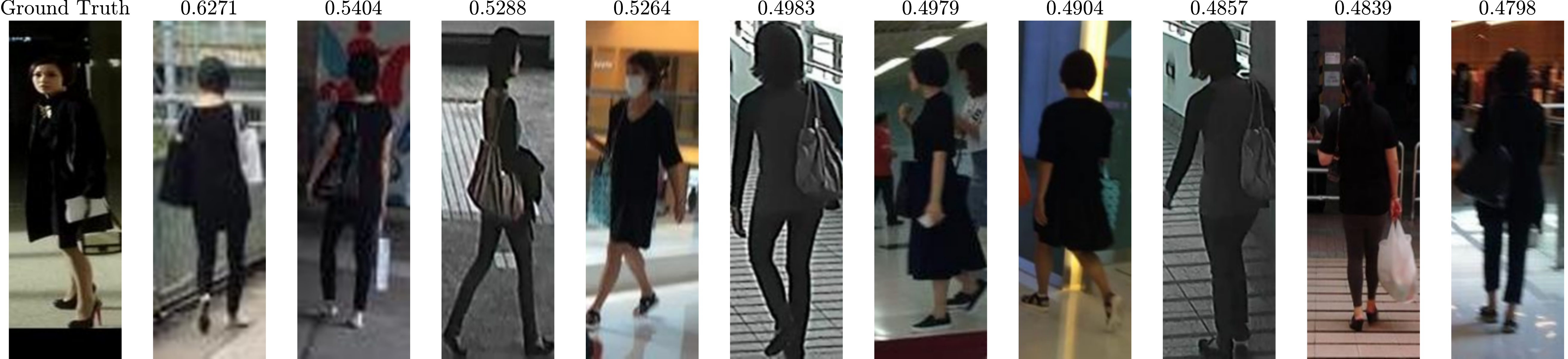}
\label{vis_2}
}
\subfigure[The \underline{woman} is wearing a \underline{thigh-level black coat} and \underline{tall black high heels} on her feet.]{
\includegraphics[width=\linewidth]{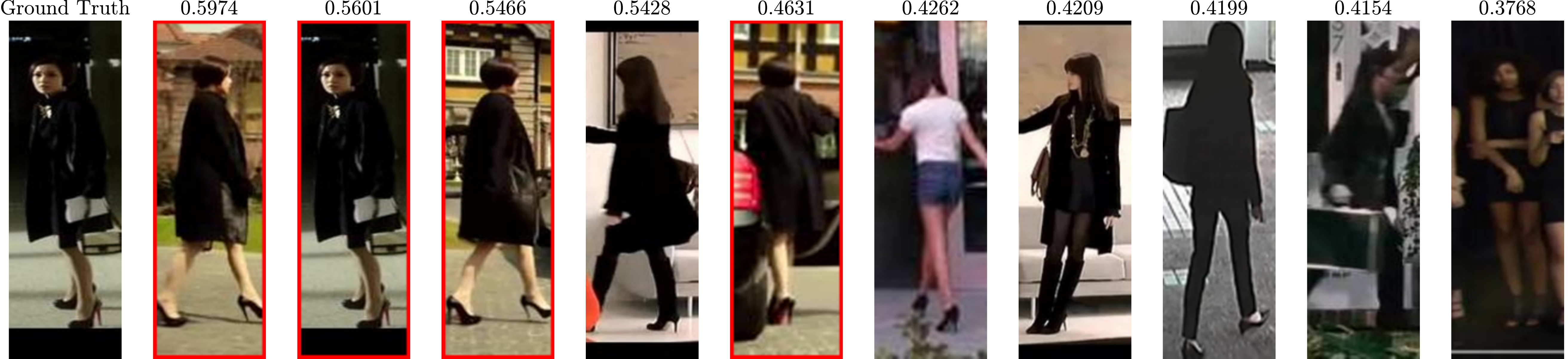}
\label{vis_3}
}
\caption{Typical retrieval results. The image with red box is the correct matching. The number on top of each image represents the predicted similarity with given text query.}
\label{vis_result}
\end{figure}

%% file: file/5-conclusions.tex
\section{Conclusion}
In this paper, we propose a novel method for text-based person search (TBPS).
Our model learns more discriminative features by using the proposed cross-modal momentum contrastive learning strategy, and
effectively transfers the knowledge learned from generic image-text pairs to compensate the data-scarce problem.
This is demonstrated by the fact that our approach clearly outperforms prior art on CUHK-PEDES, often by a big margin.

\section*{Acknowledgments}
Xiao Han appreciates Freda Shi for her helpful discussion and Kecheng Zheng for sharing details of his implementation.

%% file: file/supp_body.tex
\appendix
\section{Testing CLIP on person image classification}
As illustrated in Figure \ref{clip_error}, CLIP cannot distinguish between fine-grained features when we test it on zero-shot fine-grained person image classification. In this toy experiment, each label consists of one color and one garment, \eg, "the color of her bag is orange". However, CLIP tends to predict almost all mentioned garments as orange, demonstrating that it cannot focus on fine-grained information well.
\input{figure/clip_error}

An intuitive explanation to this phenomenon is that CLIP is trained to distinguish different visual classes using text, which is limited for intra-class discrimination in TBPS. However, as it can distinguish different visual classes using a single word (representing the class label), it thus learns an informative cross-modal representation for each word. Therefore, we use the text encoder of CLIP to embed words in each sentence, and then append a Bi-GRU to contextualize them.

\section{Training details}

\subsection{Modified ResNet101 for CLIP image encoder} 
According to CLIP \cite{radford2021clip}, this modified ResNet has three improvements over the vanilla version: (1) There are now 3 stem convolutions as opposed to 1 with an average pooling instead of max pooling. (2) It performs anti-aliasing strided convolutions, where an average pooling is prepended to convolutions with stride greater than 1. (3) The final pooling layer is a self-attention pooling instead of a global average pooling.

\subsection{Pseudocode of CM-MoCo in Pytorch-style} To better demonstrate our proposed CM-MoCo, we provide a pseudocode in Pytorch-style as following.

\definecolor{codeblue}{rgb}{0.25,0.5,0.5}
\lstset{
  backgroundcolor=\color{white},
  basicstyle=\fontsize{8.5pt}{8.5pt}\ttfamily\selectfont,
  columns=fullflexible,
  breaklines=true,
  captionpos=b,
  commentstyle=\fontsize{8.5pt}{8.5pt}\color{codeblue},
  keywordstyle=\fontsize{8.5pt}{8.5pt},
  frame=tbrl,
}
\begin{lstlisting}[language=python]
  # f_v_q, f_v_k: encoder networks for visual query and key
  # f_t_q, f_t_k: encoder networks for textual query and key
  # queue_t, queue_v, queue_id: queues to store K keys
  # m: momentum (0.999)
  # t: temperature (0.07)
  # ----------------------------------------------------------
  # bmm: batch matrix multiplication
  # mm: matrix multiplication
  # cat: concatenation
  # complement: get complement set
  
  f_v_k.params, f_t_k.params = f_v_q.params, f_t_q.params  # initialize
  
  for v, t, pid in loader:  # load a batch data with B samples
    v_q = f_v_q.forward(v)  # visual queries: BxD
    t_q = f_t_q.forward(t)  # textual queries: BxD
    v_k = f_v_k.forward(v)  # visual keys: BxD
    t_k = f_t_k.forward(t)  # textual keys: BxD
    
    # stop gradients for keys
    v_k, t_k = v_k.detach(), t_k.detach()
    
    # positive logits: Bx1
    v_pos = bmm(v_q.view(B, 1, D), t_k.view(B, D, 1))
    t_pos = bmm(t_q.view(B, 1, D), v_k.view(B, D, 1))
    
    # get P indexes of the positive instances in the queue,
    # whose identity exist in the current batch
    pos_idx = queue_id.expand(B, K).eq(pid.unsqueeze(-1)).nonzero()[:, 1]
    neg_idx = arange(K).complement(pos_idx)  # negative indexes: K-P
    
    # negative logits: Bx(K-P)
    v_neg = mm(v_q.view(B, D), queue_t.view(D, K))[:, neg_idx]
    t_neg = mm(t_q.view(B, D), queue_v.view(D, K))[:, neg_idx]
    
    # logits: Bx(1+K-P)
    logits_v = cat([v_pos, v_neg], dim=1)
    logits_t = cat([t_pos, t_neg], dim=1)
    
    # contrastive loss
    labels = zeros(B)  # positives are the 0-th
    loss = CrossEntropyLoss(logits_v / t, labels) \ 
         + CrossEntropyLoss(logits_t / t, labels)
    
    # gradient update
    loss.backward()
    
    # momentum update
    f_v_k.params = m * f_v_k.params + (1 - m) * f_v_k.params
    f_t_k.params = m * f_t_k.params + (1 - m) * f_t_k.params
    
    # update queues
    enqueue(queue_v, v_k)  # enqueue the current batch
    enqueue(queue_t, t_k)
    enqueue(queue_id, pid)
    dequeue(queue_v)       # dequeue the earliest batch
    dequeue(queue_t)
    dequeue(queue_id)
\end{lstlisting}

\vspace*{-4mm}
\subsection{Alignment loss} We discard the widely used CMPM loss \cite{zhang2018cmpc} and utilize the logistic-based contrastive loss proposed in ViTAA \cite{wang2020vitaa} as our cross-modal alignment loss. Particularly, for the visual side, given an image q-feature $\mathbf{V}_i^q$ and a batch of text q-features $\mathbf{T}_q$, the cross-modal cosine similarity $\mathbf{S}_i$ is calculated by $\mathbf{S}_i=\mathbf{V}_i^q\otimes\mathbf{T}_q^T$, where $\mathbf{S}_i \in \mathbb{R}^B$ and $\otimes$ denotes matrix multiplication. For the textual side, the calculation is identical and implemented by multiplying the alignment loss by 2. The alignment loss is finally defined as following formula \ref{align_loss}, where $\mathbf{S}_i^{+}/\mathbf{S}_i^{-}$, $\tau_{p}/\tau_{n}$ and $\alpha/\beta$ denotes the similarity, temperature and absolute margin for positive/negative pairs, respectively.
\begin{equation}
\label{align_loss}
\small
\mathcal{L}_{align}=\frac{2}{B} \sum_{i=1}^{B}\left\{\log \left[1+e^{-\tau_{p}\left(\mathbf{S}_{i}^{+}-\alpha\right)}\right]+\log \left[1+e^{\tau_{n}\left(\mathbf{S}_{i}^{-}-\beta\right)}\right]\right\}.
\end{equation}

Our consideration on the alignment loss is two folds: (1) Unlike triplet loss only considers the relative distances or CMPM \cite{zhang2018cmpc} adopts KL divergence to associate the representations across different modalities in a batch, our alignment loss considers both relative and absolute distances between positive and negative pairs; (2) $\tau_{p}$ and $\tau_{n}$ can adjust the slope of the back propagation gradient according to \ref{gradient}, which will assign higher weights to more informative samples and then lower the risk of slow convergence or even model degeneration. 
\begin{equation}
\label{gradient}
\frac{\partial \mathcal{L}_{align}}{\partial \mathbf{S}_{i}^{+}}=\frac{-\tau_{p}}{1+e^{\tau_{p}\left(\mathbf{S}_{i}^{+}-\alpha\right)}},\quad \frac{\partial \mathcal{L}_{align}}{\partial \mathbf{S}_{i}^{-}}=\frac{\tau_{n}}{1+e^{\tau_{n}\left(\beta-\mathbf{S}_{i}^{-}\right)}}.
\end{equation}

\subsection{Identity loss} We also regard identity classification with $N$ labels as an auxiliary task. Cross entropy loss \ref{id_loss} is adopted here to assist the learning of instance discriminative features. $\mathbf{W} \in \mathbb{R}^{D\times N}$ denotes a shared projection matrix following visual and textual streams. Because
person identities in the testing set do not appear in the training set, it is of importance to prevent the model from overfitting to the training identities. To this end, we replace the original one-hot label of each identity with a softer version by means of Label Smooth (LS) \cite{szegedy2016ls, luo2019strong}  with the smooth factor $\epsilon=0.1$.
\begin{equation}
\label{id_loss}
\small
\mathcal{L}_{id}=\frac{1}{B} \sum_{i=1}^{B}-\log \left(\frac{e^{ \mathbf{W}_{id_{i}}^{\top} \mathbf{V}_{i}^q}}{\sum_{j}^N e^{\mathbf{W}_{j}^{\top} \mathbf{V}_{i}^q}}\right) + \frac{1}{B} \sum_{i=1}^{B}-\log \left(\frac{e^{ \mathbf{W}_{id_{i}}^{\top} \mathbf{T}_{i}^q}}{\sum_{j}^N e^{\mathbf{W}_{j}^{\top} \mathbf{T}_{i}^q}}\right).
\end{equation}

\subsection{Rerank post-processing} In the inference stage, only two q-encoders are used. We also incorporate the multimodal $k$-reciprocal rerank algorithm proposed in NAFS \cite{gao2021contextual} into our post-processing to further improve the performance. For the text-to-image task, the initial ranking list is obtained by sorting the cross-modal cosine similarity calculated by the text query $t$ and each gallery image $v$. For each image $v$, the $k$-nearest neighboring images are obtained with the visual unimodal cosine similarity, denoted as $N_{i2i}(v, k)$. Similarly, the nearest image neighbors for the textual query $N_{t2i}(t, k)$ are obtained based on the cross-modal similarity. Finally, the pair-wise rerank similarity $D_J(v, t)$ \ref{rerank} is calculated by Jaccard Distance and added to the original cosine similarity with a weight of 0.05. For the image-to-text task, we extend this formula in a symmetrical manner to obtain $D_J(t, v)$.
\begin{equation}
\label{rerank}
D_{J}(v, t)=1-\frac{N_{i 2 i}(v, k) \bigcap N_{t 2 i}(t, k)}{N_{i 2 i}(v, k) \bigcup N_{t 2 i}(t, k)},\quad D_{J}(t, v)=1-\frac{N_{t 2 t}(t, k) \bigcap N_{i 2 t}(v, k)}{N_{t 2 t}(t, k) \bigcup N_{i 2 t}(v, k)}.
\end{equation}

\section{More evaluation results}
\subsection{The model pre-trained on MSCOCO}
There are many other available models \cite{lu2019vilbert,tan2019lxmert,li2020oscar,chen2020uniter} pre-trained on large-scale generic image-text pairs \cite{lin2014coco,plummer2015flickr30k}. However, we choose the experiment settings used in VSE++ \cite{vaghri2019vsepp} to prepare our comparative experiments. Our consideration is two-fold: (1) VSE++ is designed in a two-stream manner, which guarantees a high inference speed for TBPS; (2) No detection module, \eg, Faster-RCNN \cite{ren2016faster}, is used in VSE++, leading to a more fair comparison. We change the triplet loss used in VES++ into our alignment loss and no hard example mining is used. The results of our model can be found in Table \ref{comp_coco}.

\begin{table}[h]
\resizebox{\textwidth}{!}{
\begin{tabular}{|c|c|c|c|c|c|c|c|}
\hline
\multirow{2}{*}{\textbf{Model}} & \multirow{2}{*}{\textbf{Trainset}} & \multicolumn{3}{c|}{\textbf{Image Retrieval}} & \multicolumn{3}{c|}{\textbf{Caption Retrieval}} \\ \cline{3-8} 
                                &                                    & \textbf{R@1}  & \textbf{R@5}  & \textbf{R@10} & \textbf{R@1}   & \textbf{R@5}  & \textbf{R@10}  \\ \hline \hline
VSE++ (VGG19, GRU, FT)          & RC+rV                              & 24.1          & 52.8          & 66.2          & 32.9           & 61.7          & 74.7           \\ \hline
VSE++ (ResNet152, GRU, FT)      & RC+rV                              & \second{30.3}          & \second{59.4}          & \second{72.4}          & \second{41.3}           & \second{71.1}          & \second{81.2}           \\ \hline
Ours (ResNet101, BERT)          & RC+rV                              & \first{33.2}          & \first{63.5}          & \first{75.1}          & \first{46.8}           & \first{76.3}          & \first{85.7}           \\ \hline
\end{tabular}
}
\vspace*{1mm}
\caption{Comparison between our pre-trained model and VSE++ \cite{vaghri2019vsepp} on MSCOCO \cite{lin2014coco}. All results are calculated in MSCOCO 5k test split. FT, RC and rV denote fine-tune, random crop and rest validation set, respectively. Please refer to the paper of VSE++ \cite{vaghri2019vsepp} for details.}
\label{comp_coco}
\end{table}

\vspace*{-2mm}
\subsection{Model size and retrieval efficiency}
Table \ref{efficiency} shows the comparisons of model size and retrieval efficiency between our method and the previous state of the art. In addition to the higher retrieval performance, our method also has three advantages: (1) Our architecture, no matter is built upon ResNet50 or 101, has much fewer parameters than those of other methods because of the single-scale architecture. A smaller model size leads to less GPU memory usage and faster training speed. (2) Our method has the fastest retrieval time because only global features are used during retrieval. This advantage can guarantee real-time retrieval and thus is friendly to practical deployment. (3) Our method has the least offline feature storage because we do not need to store local information and our features' embedding dimension (256) is quite smaller than that of NAFS (768) and TIPCB (2048). Small storage usage is crucial for practical cases with scaled-up data, otherwise it will increase the burden of the whole system and the cost of computing.
\begin{table}[ht]
\resizebox{\textwidth}{!}{
\begin{tabular}{|c|c|c|c|c|c|}
\hline
\multirow{2}{*}{\textbf{Model}} & \multirow{2}{*}{\textbf{Rank-1} $\uparrow$} & \multirow{2}{*}{\textbf{Params (M) $\downarrow$}} & \multirow{2}{*}{\textbf{Retrieval Time (s) $\downarrow$}} & \multicolumn{2}{c|}{\textbf{Offline Feature Storage $\downarrow$}} \\ \cline{5-6} 
                                &                                  &                                      &                                               & \textbf{Visual Side}      & \textbf{Textual Side}     \\ \hline \hline
ViTAA \cite{wang2020vitaa}                           & 54.92                            & 176.53                               & \first{0.02}                                          & \first{3MB}                       & \first{6MB}                       \\ \hline
NAFS \cite{gao2021contextual}                            & 59.36                            & 188.75                               & \second{0.07}                                              & \second{9MB}                       & \second{18MB}                      \\ \hline
TIPCB \cite{chen2021tipcb}                            & \second{63.63}                            & 184.75                               & 0.20                                              & 24MB                       & 48MB                      \\ \hline
Ours (ResNet50)                   & 61.65                            & \first{42.33}                                & \first{0.02}                                          & \first{3MB}                       & \first{6MB}                       \\ \hline
Ours (ResNet101)                  & \first{64.08}                            & \second{60.20}                                & \first{0.02}                                          & \first{3MB}                       & \first{6MB}                       \\ \hline
\end{tabular}
}
\label{efficiency}
\caption{Comparisons of model size and retrieval efficiency among ViTAA \cite{wang2020vitaa}, NAFS \cite{gao2021contextual} and our method. Retrieval time is computed by retrieving all text queries (6156) through the whole image gallery (3074) of CUHK-PEDES test set \cite{li2017person}.}
\end{table}

\vspace*{-4mm}
\section{More visualization results}
\subsection{Visualization of self-attention pooling} 
\begin{figure}[htbp]
\centering
\subfigure{
\includegraphics[width=0.23\linewidth]{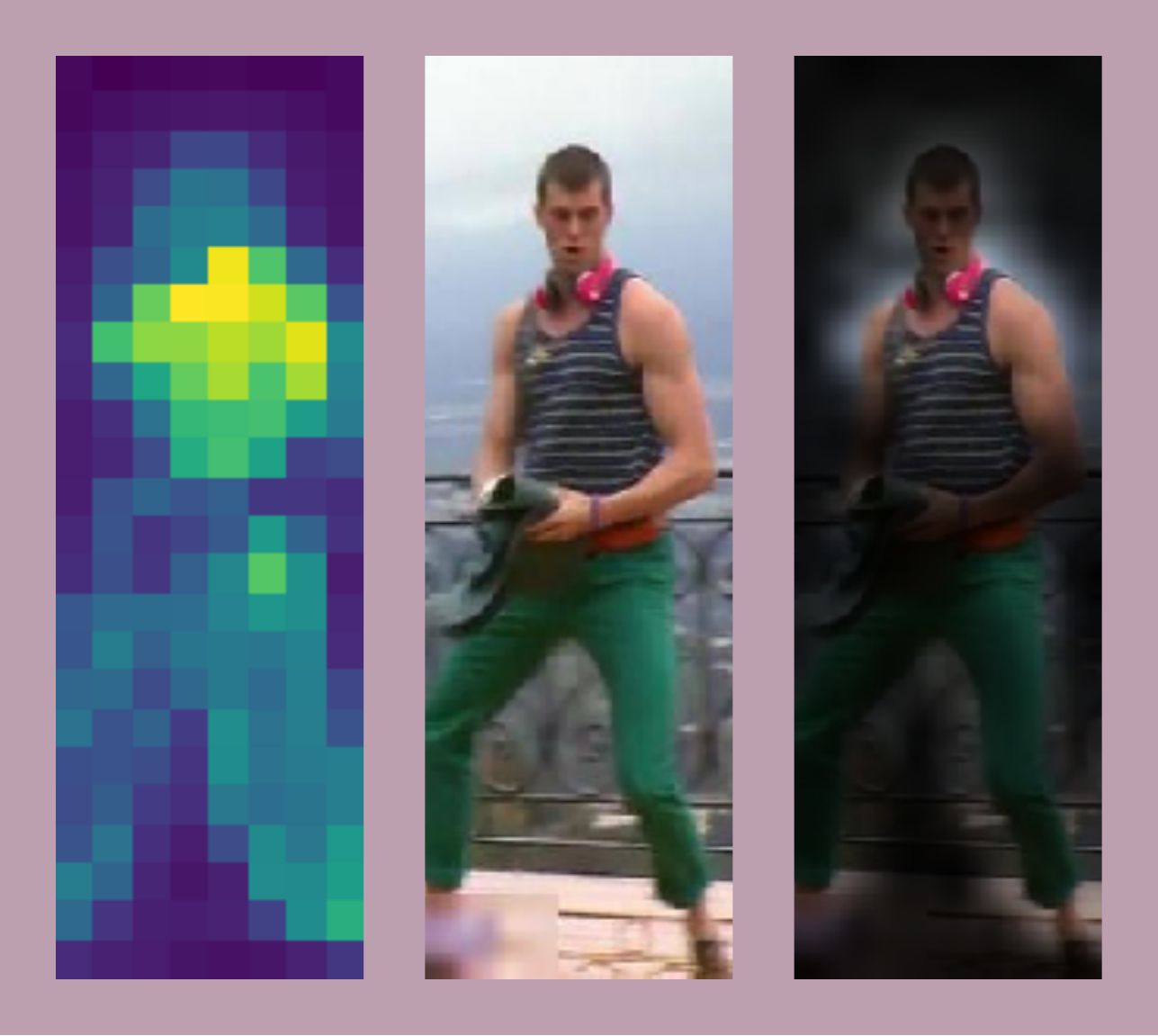}
}
\subfigure{
\includegraphics[width=0.23\linewidth]{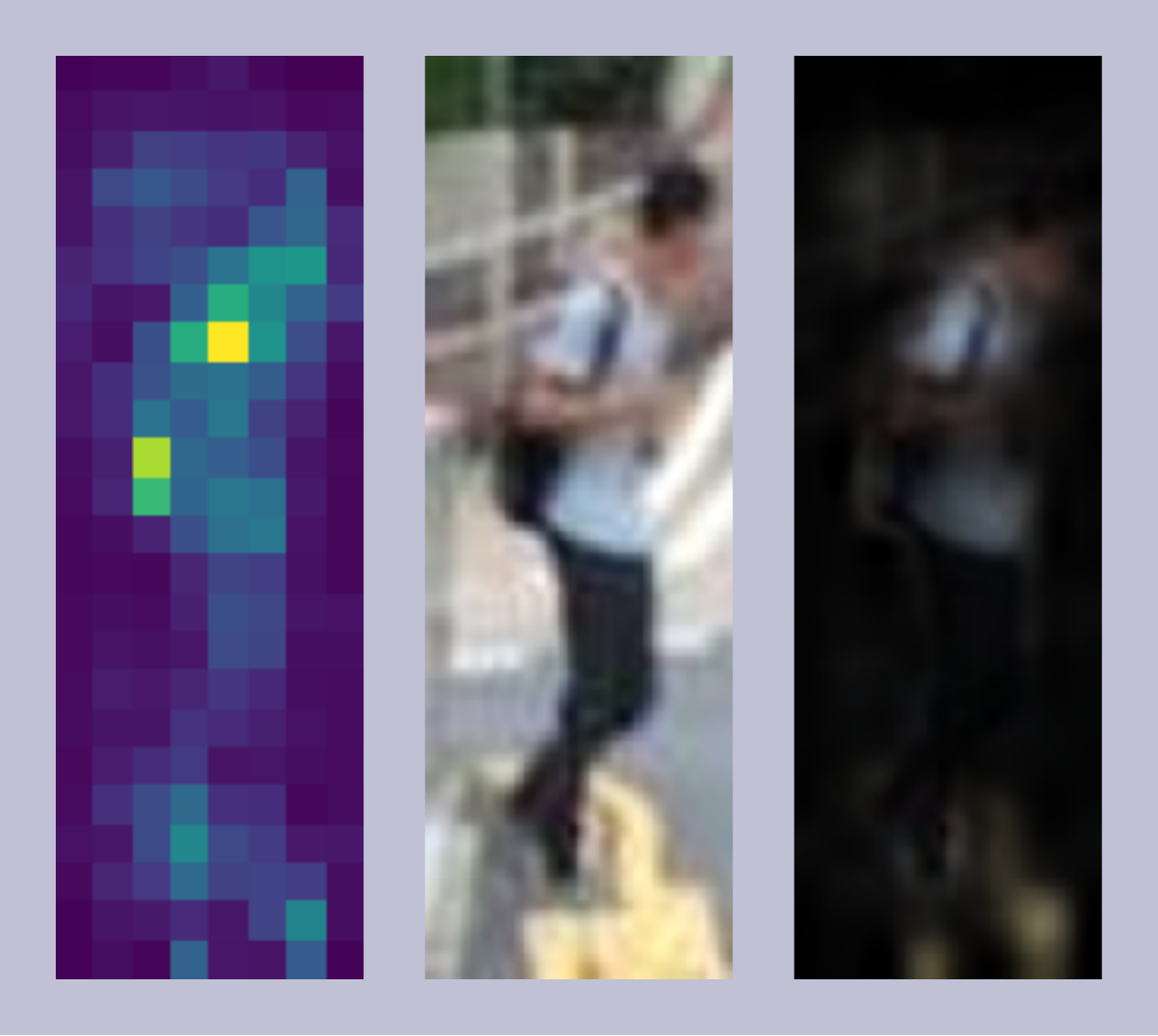}
}
\subfigure{
\includegraphics[width=0.23\linewidth]{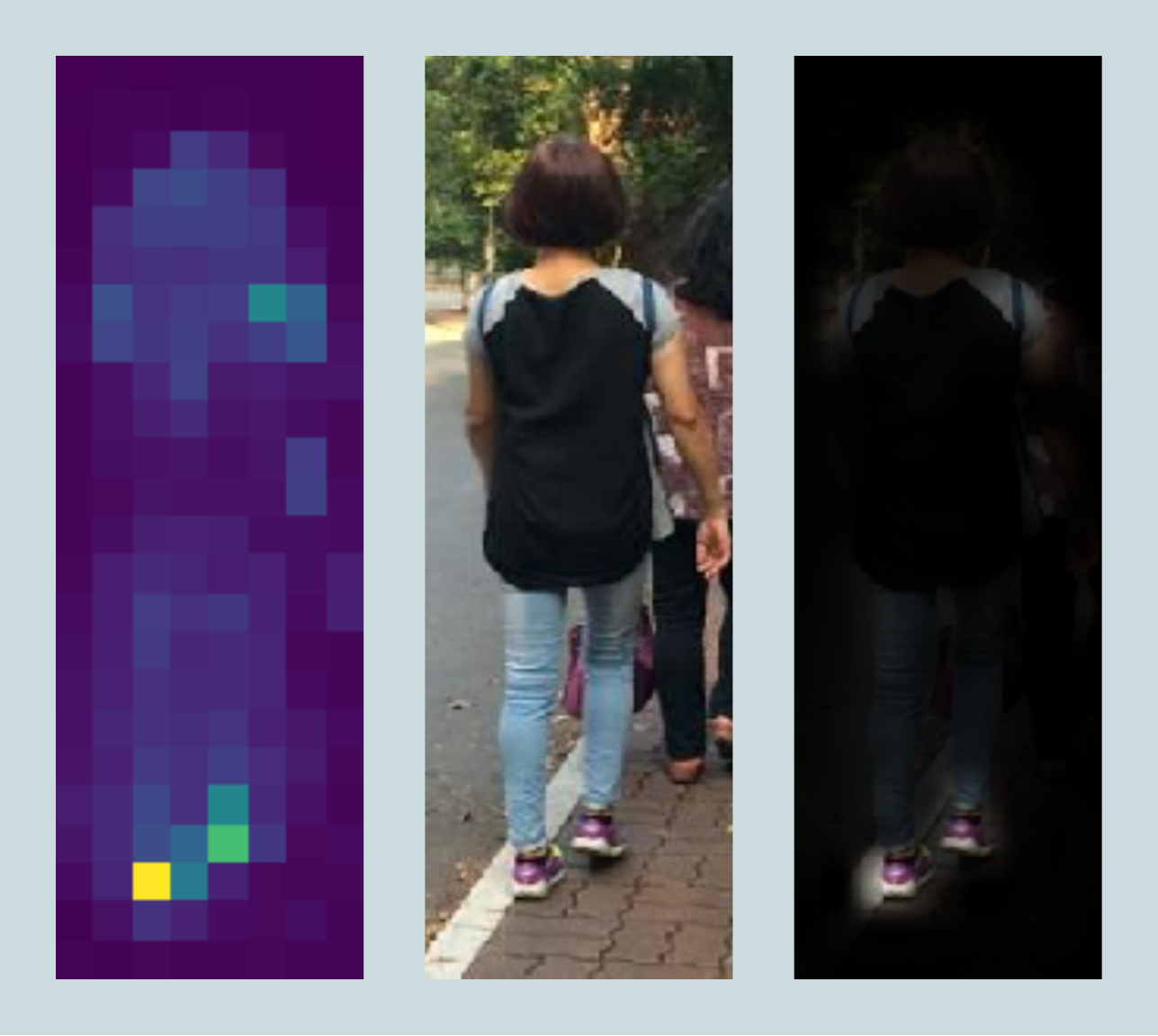}
}
\subfigure{
\includegraphics[width=0.23\linewidth]{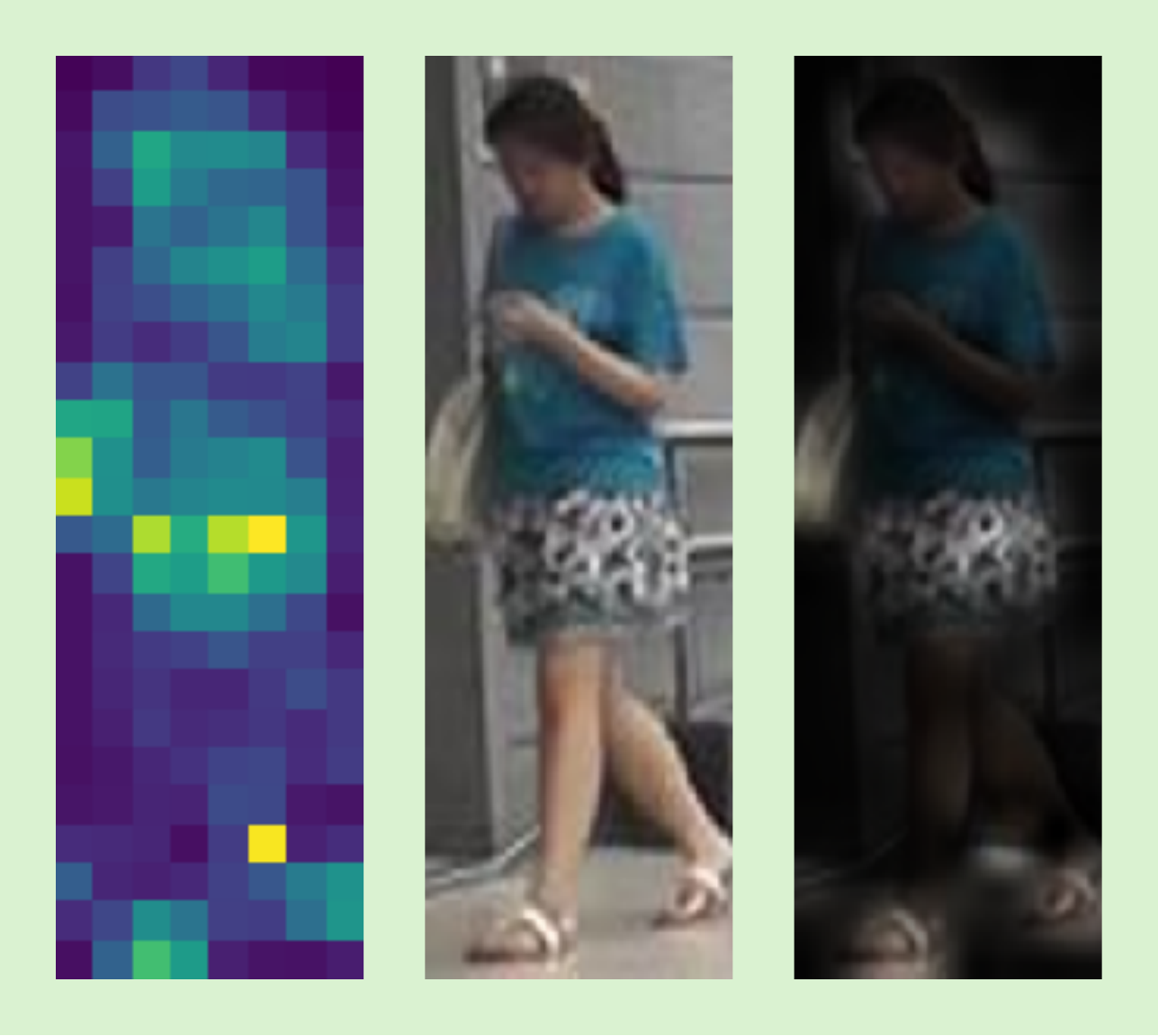}
}
\caption{Visualization of the last layer attention map calculated by [CLS] token and other patch tokens in the self-attention pooling layer of CLIP ResNet101. This figure contains attention maps, original images and images multiplied by resized attention map for four different identities randomly sampled from test set.}
\label{vis_sa}
\end{figure}

Figure \ref{vis_sa} visualizes the learned attention weight in the self-attention pooling layer of CLIP Image Encoder (ResNet101 version). We can conclude that the visual stream is capable of learning the salient parts related to the garments of a person rather than the background. This visualization further verifies that the model has the ability to learn reasonable features even without the help of multi-scale information.

\subsection{More visualized retrieval results} 
\begin{figure}[htbp]
\centering
\subfigure{
\includegraphics[width=0.95\linewidth]{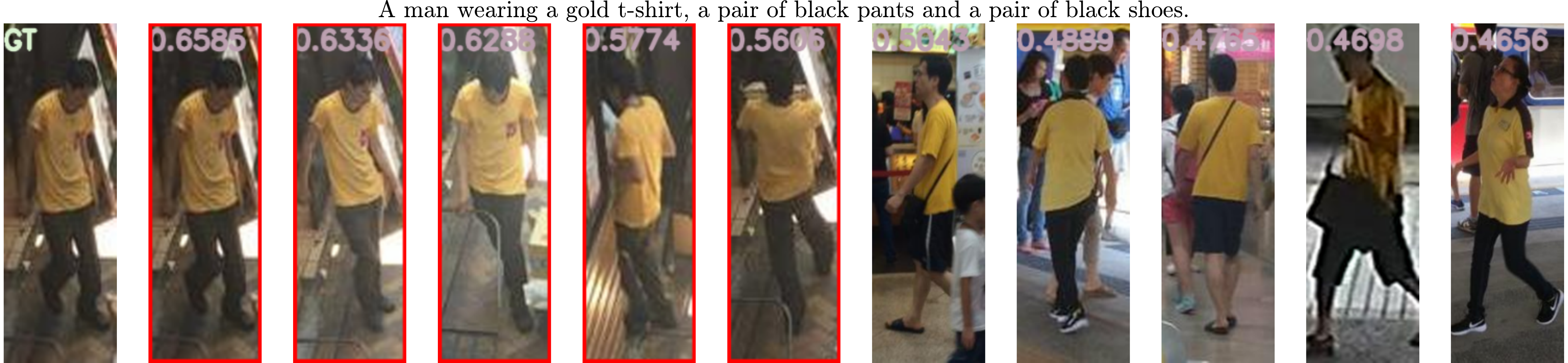}
}
\subfigure{
\includegraphics[width=0.95\linewidth]{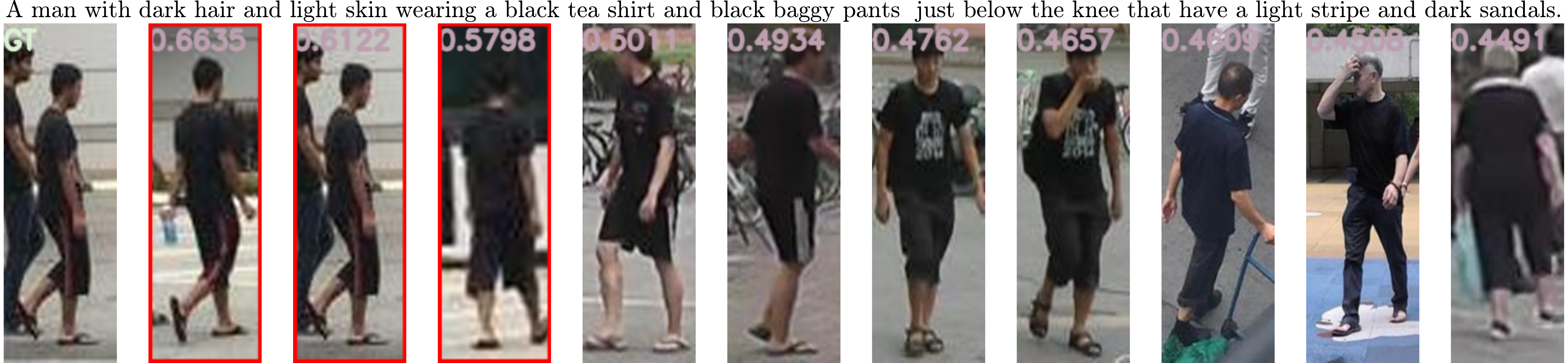}
}
\subfigure{
\includegraphics[width=0.95\linewidth]{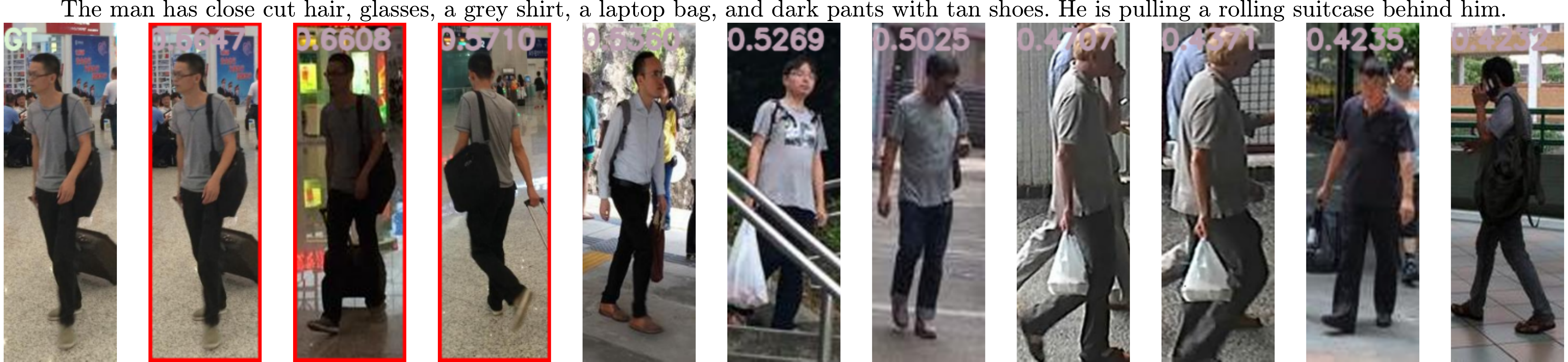}
}
\subfigure{
\includegraphics[width=0.95\linewidth]{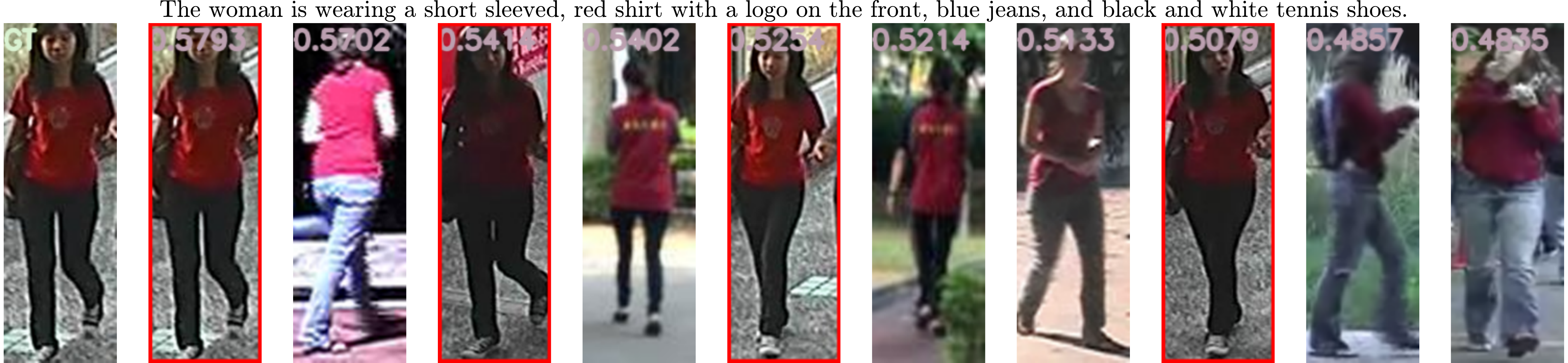}
}
\subfigure{
\includegraphics[width=0.95\linewidth]{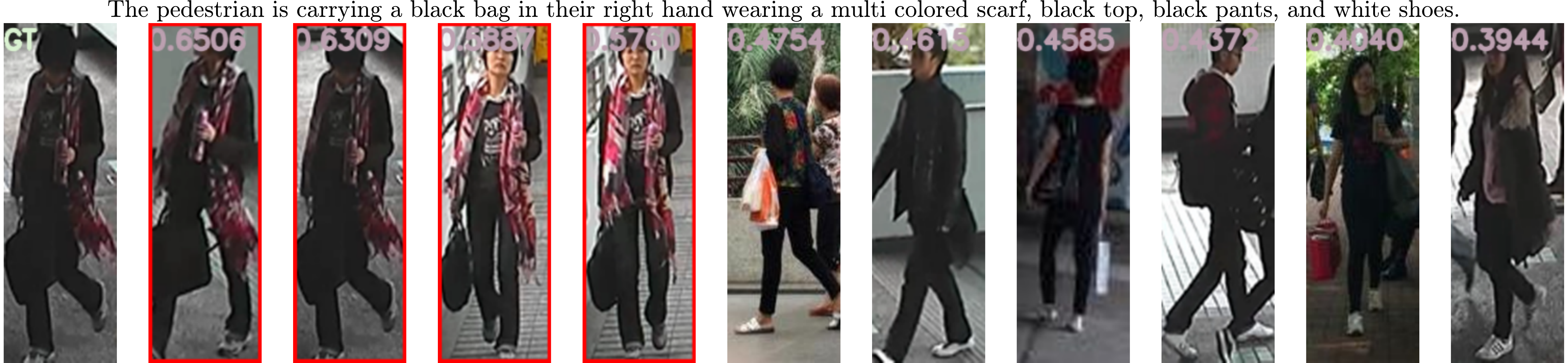}
}
\subfigure{
\includegraphics[width=0.95\linewidth]{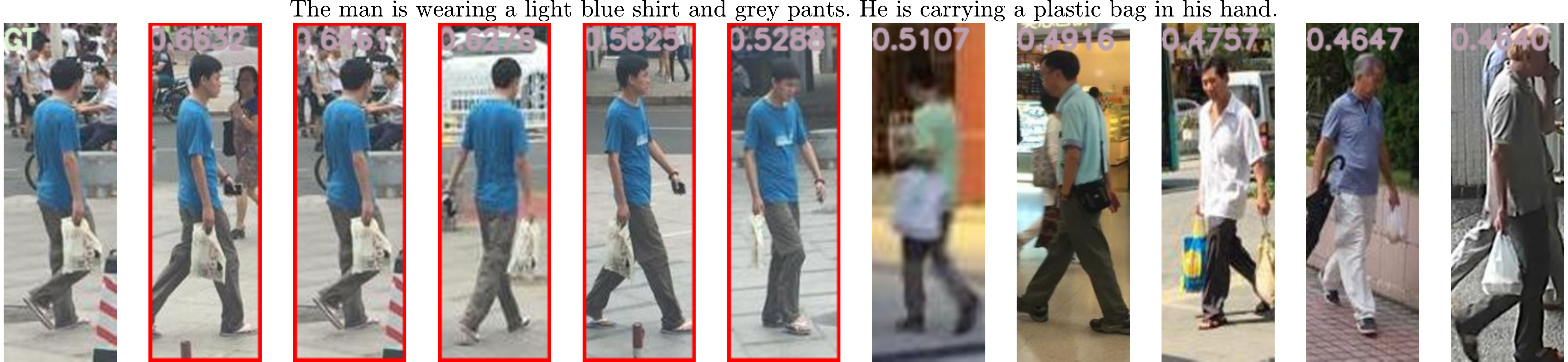}
}
\caption{Typical successful cases of retrieval results.}
\label{success_cases}
\end{figure}

\begin{figure}[htbp]
\centering
\subfigure{
\includegraphics[width=0.95\linewidth]{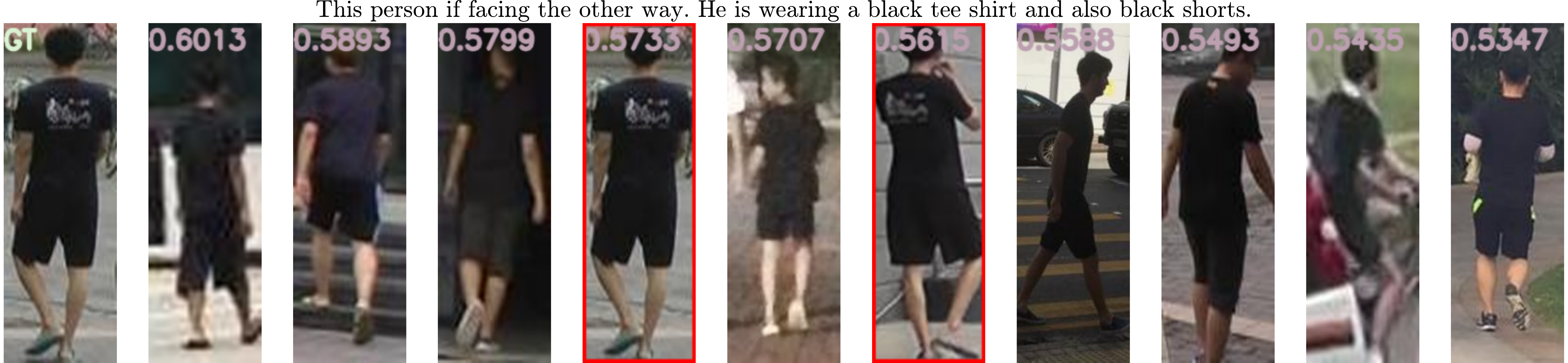}
}
\subfigure{
\includegraphics[width=0.95\linewidth]{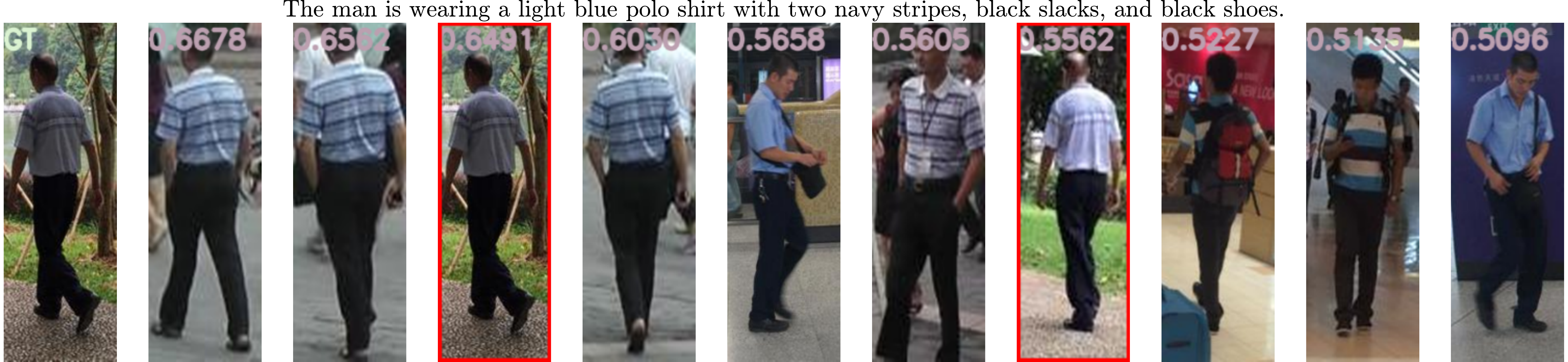}
}
\subfigure{
\includegraphics[width=0.95\linewidth]{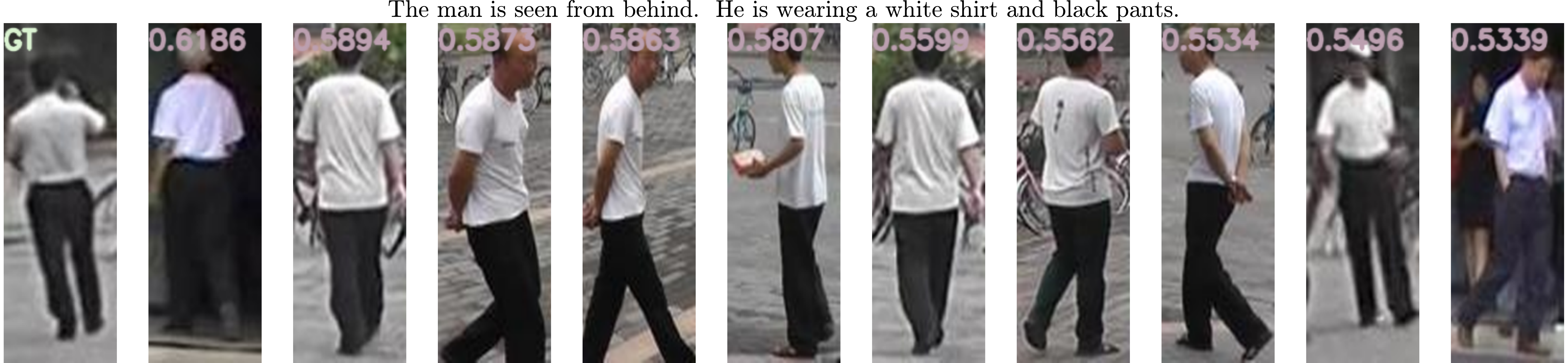}
}
\subfigure{
\includegraphics[width=0.95\linewidth]{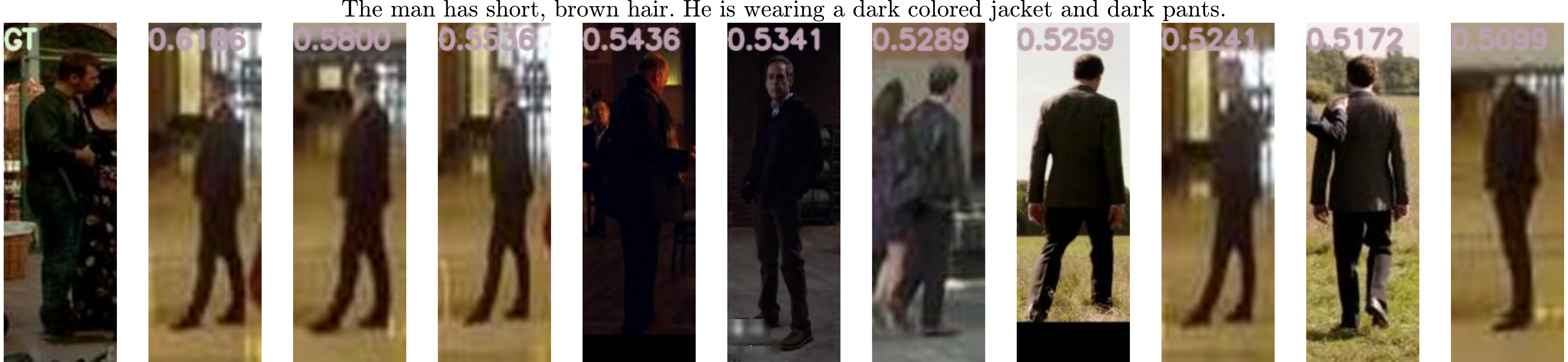}
}
\subfigure{
\includegraphics[width=0.95\linewidth]{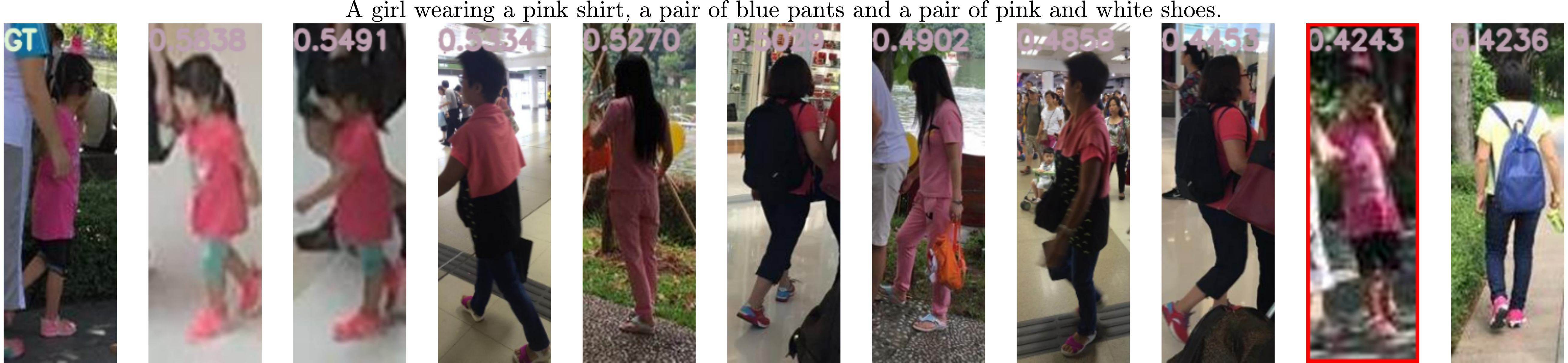}
}
\subfigure{
\includegraphics[width=0.95\linewidth]{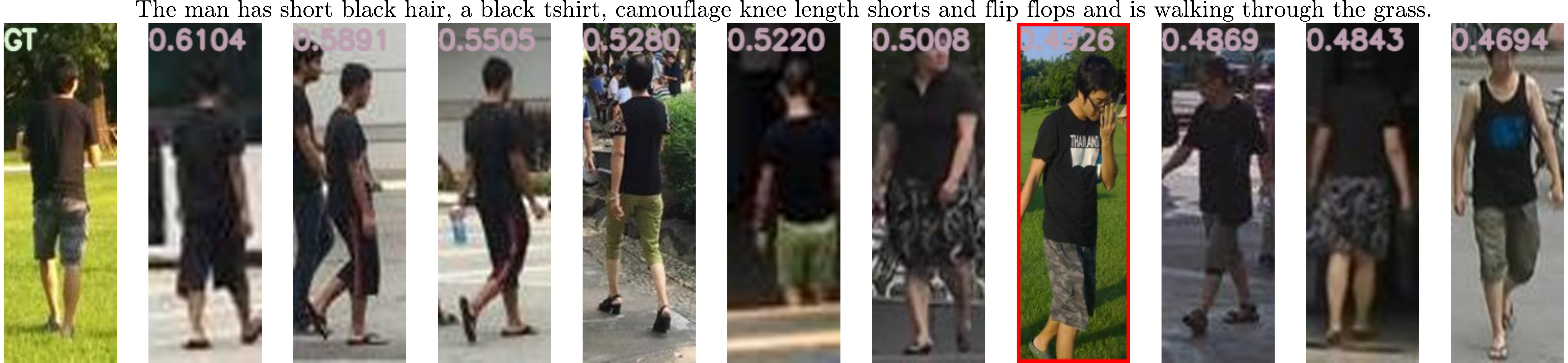}
}
\caption{Typical failure cases of retrieval results.}
\label{failure_cases}
\end{figure}

We visualize several typical successful and failure cases of our retrieval results in Figure \ref{success_cases} and \ref{failure_cases}, respectively. It is apparent that this failure cases are due to the ambiguity in the images or the 
pragmatic vagueness in the sentences. The predictions of our model are reasonable, and the more specific the search sentence is, the better our search results will be.

%% file: figure/clip_error.tex
\begin{figure}[h]
\centering
\includegraphics[width=\linewidth]{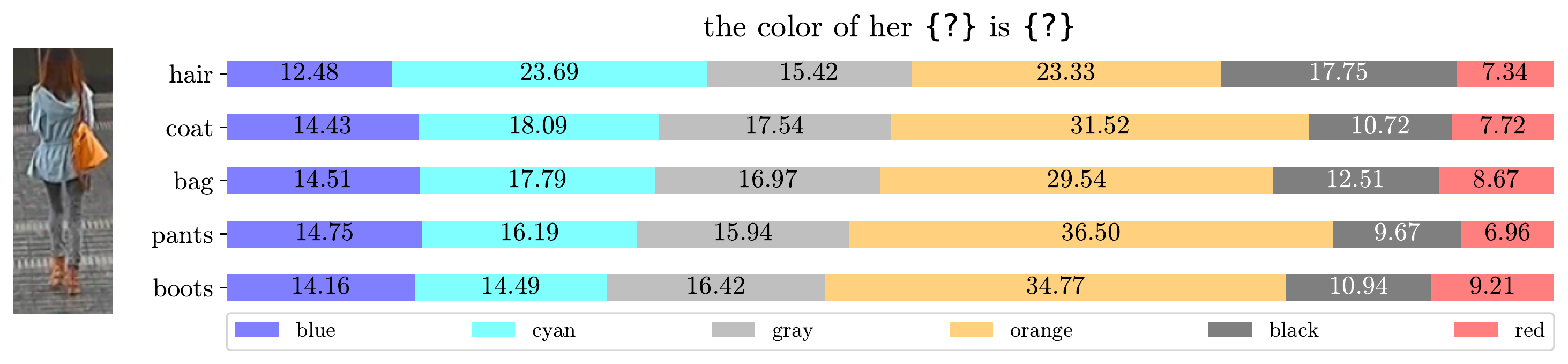}
\caption{Visualization of the probabilities predicted by CLIP \cite{radford2021clip} for fine-grained zero-shot person image classification.}
\label{clip_error}
\end{figure}